\documentclass[journal]{IEEEtran}
\usepackage{cite}
\ifCLASSINFOpdf
  \usepackage[pdftex]{graphicx}
  \DeclareGraphicsExtensions{.pdf,.png}
\else
\fi
\usepackage[cmex10]{amsmath}
\usepackage{amssymb}

\ifCLASSOPTIONcompsoc
  \usepackage[caption=false,font=normalsize,labelfont=sf,textfont=sf]{subfig}
\else
  \usepackage[caption=false,font=footnotesize]{subfig}
\fi

\usepackage{color}
\newcommand{\Rm}{\ensuremath{R^M}}
\newcommand{\Dm}{\ensuremath{D^M}}
\newcommand{\upq}{\ensuremath{u_{f_p,f_q}}}
\newcommand{\Rerr}{\ensuremath{\mathcal{R}}}
\newcommand{\tr}{\ensuremath{\text{tr}}}

\begin{document}
%\title{Bare Demo of IEEEtran.cls for Journals}
\title{Learning Discriminative Multilevel Structured Dictionaries for Supervised Image Classification}

\author{J\'er\'emy~Aghaei~Mazaheri,
        Elif~Vural,
        Claude Labit,
        and~Christine~Guillemot% <-this % stops a space
\thanks{J. Aghaei Mazaheri and E. Vural were with Institut National de Recherche en Informatique et Automatique (INRIA), Rennes, France. E. Vural is now with the Deparment of Electrical and Electronics Engineering at METU, Ankara, Turkey. C. Labit and C. Guillemot are with Institut National de Recherche en Informatique et Automatique (INRIA), Rennes, France.}
}
\maketitle

\begin{abstract}

%Sparse representation has become a popular research topic for many signal processing applications like denoising, super-resolution, inpainting, compression or classification. It has often been coupled with dictionary learning to adapt the dictionary on which the representation is done to specific data, making the reconstruction more efficient. In the application of supervised classification, the dictionaries have also been made discriminative during the learning to be better adapted to the classification issue. This article proposes to learn structured dictionaries, for their learning ability due to the number of dictionaries they contain, and to make them discriminant for classification. The reconstruction errors computed on each dictionary structure, for each class, are then used by a combination of two smoothing steps aiming to clean the segmentation of a picture into labels. Applied to a common set of texture images, our supervised classification method shows competitive results with the state of the art. Finally, the learning dataset is enriched to deal with some over-exposure problems.

Sparse representations using overcomplete dictionaries have proved to be a powerful tool in many signal processing applications such as denoising, super-resolution, inpainting, compression or classification. The sparsity of the representation very much depends on how well the dictionary is adapted to the data at hand. In this paper, we propose a method for learning structured multilevel dictionaries with discriminative constraints to make them well suited for the supervised pixelwise classification of images. A multilevel tree-structured discriminative dictionary is learnt for each class, with a learning objective concerning the reconstruction errors of the image patches around the pixels over each class-representative dictionary. After the initial assignment of the class labels to image pixels based on their sparse representations over the learnt dictionaries, the final classification is achieved by smoothing the label image with a graph cut method and an erosion method. Applied to a common set of texture images, our supervised classification method shows competitive results with the state of the art.

\end{abstract}


% \fi
\IEEEpeerreviewmaketitle

\section{Introduction}
%\IEEEPARstart{A}{demo} file is .... Introduction begins...

%Sparse representations
\IEEEPARstart{S}{parse} representations have become popular in several applications of signal, image and video processing, such as denoising \cite{elad_image_2006, dong_sparsity-based_2011}, super-resolution, inpainting, compression \cite{figueras_i_ventura_low-rate_2006, sezer_sparse_2008, bryt_compression_2008, zepeda_image_2011} or classification. While it was common to analyze and reconstruct signals based on representations over predefined bases such as wavelets and DCT, research in the recent years has shown that learning overcomplete  dictionaries adapted to the structure of the treated signals can significantly improve the representation quality. Observing that learning redundant dictionaries from collections of data samples under sparsity priors leads to models that fit and approximate well the characteristics of signals \cite{aharon_k-svd:_2006}, \cite{engan_method_1999}, the learning of dictionaries in a supervised setting for the discrimination of different classes of signals has also become a popular research problem \cite{jiang_label_2013}. In this work, we propose a method to learn multilevel structured dictionaries with high discrimination capability for the problem of pixelwise image classification.

%used to be reconstructed using predefined dictionaries, with the assumption that a given signal admits a sparse decomposition in frequently used bases (wavelets, DCT, ...), i

%, to obtain the approximation $Y \approx DX$.

%Dictionary learning for supervised classification (several methods)
We consider a supervised classification setting where the classes are known and exemplars are available for each class. In particular, we are interested in image classification problems with a large amount of variability between data samples of the same class,  resulting from e.g., dominant presence of irregular high-frequency content in the image classes, or multiple subcategories within the same image class with little resemblance between them. Some example applications could be texture classification problems where the considered image texture classes are rich in high-frequency content with little correlation between several patterns belonging to the same class differing by shifts, orientation differences, etc.; or remote sensing satellite images with high variability within the same image class (e.g., the ``city'' class containing both smooth image regions corresponding to flat areas such as parks and rivers; and regions rich in texture corresponding to populated urban areas with buildings and streets).

In this setting, we consider the problem of learning a discriminative dictionary model for each class. In order to handle the large variability or the presence of multiple subcategories of patterns in each image class, we propose to use multilevel dictionaries having a tree-like structure. In the proposed setting, the overall class-representative dictionary consists of subdictionaries residing at multiple levels, such that each subdictionary in a level originates from a certain atom of a subdictionary in the preceding level. The representation of an image patch in a multilevel dictionary is simply computed by tracing down the branches, i.e., first choosing an atom in the first-level subdictionary, then selecting an atom from the second-level subdictionary corresponding to the first atom, and similarly descending until the desired sparsity level is attained. The patches of test images are classified with respect to their reconstruction errors over each class-representative multilevel dictionary.

\begin{figure}[!t]
\centering
\includegraphics[height=5cm]{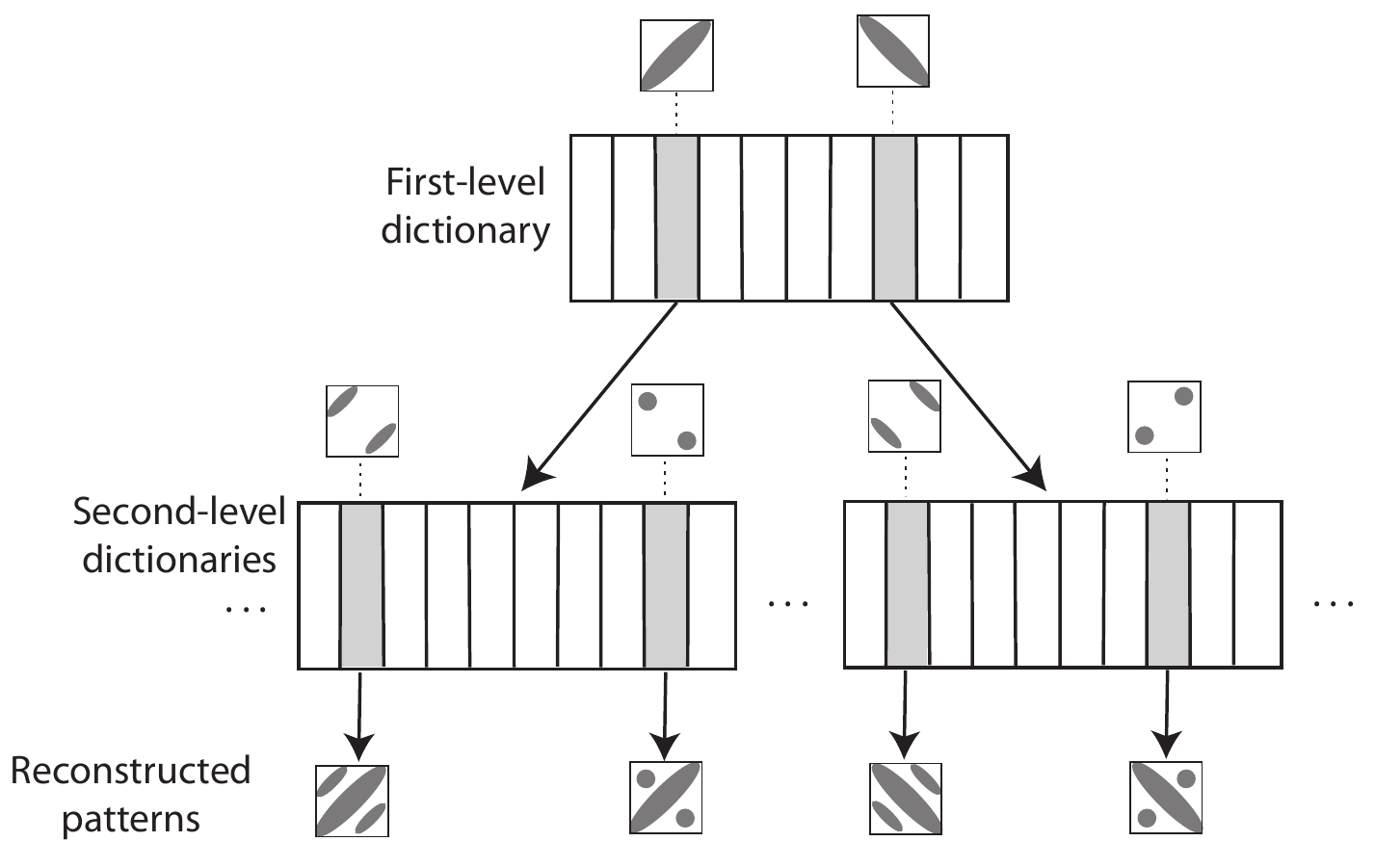}
\caption{Illustration of the proposed structured dictionaries for two levels. The atoms in the first level capture the main characteristics of different types of patterns in the same class. Each second-level dictionary originates from a different first-level atom. Second-level atoms learn the details of the patterns selecting the corresponding first-level atom.}
\label{fig:multilevel_dl_illus}
\vspace{-0.5cm}
\end{figure}

Such a multilevel dictionary structure is particularly suitable for the considered image classification problem with high intra-class variability. In a setting with various patterns of little resemblance in the same class, the atoms in upper-level subdictionaries capture the main characteristics of the patterns such as orientation, so that dissimilar patterns are represented with different atoms in these early levels. The lower-level subdictionaries originating from different atoms in upper levels are then particularly adapted to the structures of the different types of patterns present in the class and learn the fine details of these patterns. The representation of signals with the proposed multilevel structured dictionaries is illustrated in Figure \ref{fig:multilevel_dl_illus}.

%The lower-level subdictionaries stemming from different atoms in upper levels are particularly adapted to the structures of different types of patterns in the same class. While the atoms in the upper levels capture the low-level features of different patterns such as orientation, atoms in the lower levels learn the fine details of the patterns. The representation of signals with the proposed multilevel structured dictionaries is illustrated in Figure \ref{}.

%The special tree structure then allows the learning of models that are aware of the dissimilarities within the same class, such that 

Many methods in the literature use sparse representations and dictionary learning for the problem of supervised classification \cite{jiang_label_2013}, \cite{mairal_discriminative_2008}, \cite{mairal_supervised_2008}, \cite{zhang_discriminative_2010}. Using the known labels of training data, these methods learn one or several dictionaries to allow the classification of test images based on their sparse representations in the learnt dictionaries. Although the traditional single-level flat dictionaries used typically in supervised dictionary learning can learn the main characteristics of different classes via sparse representations, the main philosophy of these methods is to tune the atoms to fit well the common features in the same class while pushing them away  from the features of the other classes. While such methods give quite impressive results in applications such as face recognition with rather small variability within the same class, their performance may degrade in problems with large intra-class variability. On the other hand, the multilevel dictionary structures proposed in our method have a high learning capacity that explicitly and efficiently takes account of possible intra-class variations.

%Our paper (par rapport à la biblio)
%Our work, in the continuity of the work of Mairal et al. \cite{mairal_discriminative_2008}, focuses on learning one discriminative dictionary per class but using structured dictionaries. Indeed, we want to use the learning capacity of the structured dictionaries in the context of supervised classification, while making them discriminative to be better adapted to the classification problem. 

We propose to learn the class-representative multilevel dictionaries in a sequential way, by optimizing each atom of a subdictionary with respect to a discriminative learning objective. Our objective function seeks to update each atom in order to fit the residuals of the signals from the same class using that atom, while increasing the reconstruction error of the signals from other classes when represented with that atom. We first train the dictionaries with image patches from a set of known classes. Then, for the image patches centered around each pixel of a test picture, we compute the reconstruction errors over the learnt class-representative dictionaries. Finally, the label image is obtained by applying a combination of two smoothing methods: a label expansion algorithm based on a graph cut, and an erosion algorithm, which both use the information of the reconstruction errors of the patches over the learnt dictionaries. We evaluate our method with experiments on several texture classification problems. The experimental results show that our method gives competitive results with the state of the art. %The learning dataset is finally enriched to deal with some over-exposure problems.

%Plan of the paper
In Section \ref{sec:related_work}, we give an overview of the related work. Section  \ref{sec:proposed_dl_method} presents the proposed method for learning supervised dictionaries with a multilevel adaptive structure, together with a description of the classification algorithm based on these learnt structured dictionaries. In Section \ref{sec:smoothing}, we describe the smoothing steps applied for improving the label estimates. We present the experimental results on texture images in Section \ref{sec:exp_results}, and Section \ref{sec:conclusion} concludes the paper.

%The smoothing steps, making the classification of a picture into labels cleaner, are detailed in Section  Section \ref{sec:exp_results} is devoted to the experiments conducted on texture images, and Section \ref{sec:conclusion} concludes this article.

\section{Related work}
\label{sec:related_work}

We now briefly overview some related works on sparse representations and dictionary learning.

\subsection{Sparse representations and unsupervised dictionary learning}

Sparse representations consist in representing a signal $y\in \mathbb{R}^{n}$ as a linear combination of only a few columns, known as atoms, from a dictionary $D \in \mathbb{R}^{n\times K}$, under a sparsity constraint as
\begin{equation}
\label{eq:pb}
\underset{x}{\min}||y-Dx||^2_2, \; \textrm{subject to} \; ||x||_0 \leq L
\end{equation}
where $x \in \mathbb{R}^{K}$ is the coefficient vector corresponding to the sparse representation of $y$ over $D$, and $L > 0$ is the sparsity constraint, i.e., the maximum number of non-zero coefficients in $x$. The $l_0$-norm $||x||_0$ of $x$ is equal to the number of non-zero coefficients in $x$. The dictionary $D$ is composed of $K$ atoms $d_k, k=1,...,K$, that are supposed to be normalized to have unit $l_2$-norm as $||d_k||_2=1, \forall k=1,...,K$. 

The computation of the sparse approximation of a signal in \eqref{eq:pb} is an NP-hard problem and some greedy algorithms have been developed to find an approximate solution, such as the Matching Pursuit (MP) \cite{mallat_matching_1993} and the Orthogonal Matching Pursuit (OMP) \cite{pati_orthogonal_1993} algorithms, which search in each iteration the atom of the dictionary that is the most correlated with the current residual vector. Several other methods such as the Basis Pursuit algorithm \cite{chen_atomic_1998} propose to relax the optimization problem by replacing the $l_0$-norm of $x$ with its $l_1$-norm.

Many dictionary learning methods have been proposed to learn a dictionary $D \in \mathbb{R}^{n\times K}$ from a set of $N$ training vectors $Y \in \mathbb{R}^{n\times N}$ under sparsity constraints, in order to better adapt the dictionary to the data. Unsupervised dictionary methods typically solve the problem
\begin{equation}
\label{eq:pb_dico}
\underset{D,X}{\min}||Y-DX||_F^2, \; \textrm{subject to} \; ||x_i||_0 \leq L \; \forall i \; \textrm{and} \; ||d_k||_2=1 \; \forall k
\end{equation}
where $L > 0$ is the sparsity constraint applied to each column $x_i$ of $X$, i.e., the maximum number of non-zero coefficients in $x_i, i=1,...,N$, and $||.||_F$ is the Frobenius norm.

Many dictionary learning algorithms such as the Method of Optimal Directions (MOD) \cite{engan_method_1999, engan_frame_1999-1} and K-SVD \cite{aharon_k-svd:_2006}  apply an iterative optimization procedure with two major steps. The first step consists of the sparse coding of the training vectors over the fixed dictionary $D$ to compute $X$, which can be solved with pursuit algorithms; and the second step is the update of the dictionary $D$ based on the decompositions computed in the previous step. Some dictionary learning algorithms impose constraints on the dictionary, such as the Sparse K-SVD method\cite{rubinstein_double_2010}, which aims to learn a sparse dictionary, or the Non-Negative K-SVD method \cite{aharon_k-svd_2005}, which learns a non-negative dictionary. An online dictionary learning method based on stochastic approximations is proposed in \cite{mairal_online_2010}. Finally, structured multilevel dictionaries are learnt in \cite{zepeda_iteration-tuned_2010, zepeda_image_2011} based on the idea of adapting each dictionary to one iteration of the pursuit algorithm, so that atoms are sequentially selected from dictionaries at different levels by going down the branches in sparse coding. While the multilevel dictionary structure used in our method is based on the principle developed in these previous works, we focus here on the supervised learning problem for classification applications unlike these works.

\subsection{Supervised dictionary learning}
%1 global dico (4 meths)
Supervised dictionary methods aim to learn dictionaries such that sparse representations of signals over the learnt dictionaries allow an accurate estimation of their class labels. Some dictionary learning methods learn one global dictionary to represent all classes. The study in \cite{rodriguez_sparse_2008} shows the advantage of learnt dictionaries over predefined dictionaries in classification, where a dictionary is learnt with a discrimination term applied on the coefficients. A discriminative formulation with a linear and bilinear classifier applied to the sparse coefficients is employed in \cite{mairal_supervised_2008}. A discriminative version of K-SVD is presented in \cite{zhang_discriminative_2010}. A classifier is jointly learnt with the dictionary and then applied to the coefficients of a test picture to classify it. Applied to face recognition, it offers better results than the K-SVD dictionary. The problem of \cite{zhang_discriminative_2010} is extended in the Label Consistent K-SVD method \cite{jiang_learning_2011}, \cite{jiang_label_2013}. The dictionary is learnt along with a linear classifier using the sparse coefficients in order to increase the discrimination capability of the coefficients, while another term in the objective directly imposes the similarity of the sparse coefficients among the samples from the same class. The methods in  \cite{zhou_bilevel_2017} and \cite{yankelevsky_structure_2017} are based on similar formulations while they also include a graph-regularization term on the sparse coefficients. The authors of \cite{zhou_bilevel_2017} further propose to remove the sparse reconstruction term from the objective function and include it only in the constraints of the optimization problem. In the sparse decomposition of a training sample, the coefficients corresponding to other classes are suppressed with a differentiable term based on the $\ell_2$-norm in \cite{wang_crosslabel_2017}, while a graph-regularization term is also included in the objective.
%
%Each atom of the dictionary is associated with a particular label in order to increase the discrimination capability of the coefficients and a new term in the objective function forces signals from the same class to have very similar sparse decompositions.
%Meth intermediaire (1 meth)
%A dictionary composed of subdictionaries, each one associated with a class, is learnt in \cite{yang_fisher_2011}.  The sparse coefficients together with the reconstruction error obtained for each class are then used for classification. The objective function thus includes a discriminative fidelity term, assuring that the data from a given class are well represented on the global dictionary, and in particular on the corresponding subdictionary, but not on the other subdictionaries. The function also contains a sparsity constraint term on the coefficients, and a discriminative term, based on the Fisher discrimination criteria, applied on the coefficients to make them similar among a class but different between classes.
%A dictionary composed of subdictionaries, each one associated with a class, is learnt in \cite{yang_fisher_2011}.  The sparse coefficients together with the reconstruction error obtained for each class are then used for classification, with...
%
% Fisher, Bayesian
 A Fisher criterion is applied on the sparse coefficients in the learning in \cite{yang_fisher_2011}. The dictionary learning problem is formulated in a Bayesian setting in \cite{akhtar_discriminative_2016}, such that sparsity is imposed via class-dependent Bernoulli random vectors, and a classifier is trained on sparse codes. 
%SSL
A couple of other methods consider the semi-supervised dictionary learning problem. 
A linear classifier on sparse codes is learnt in \cite{wang_adaptively_2015} while the  unlabeled samples are also incorporated in the discriminative term of the learning objective, proportionally to the confidence of their label estimates.
% Soft label SSL
The authors of \cite{jian_semi_2016} emphasize that there may be overlapping features between different classes and propose to learn a global dictionary along with the corresponding soft label vectors in a graph-regularized semi-supervised learning scheme. 

%1 dico / class (3 meths)
Some other methods learn one dictionary per class and classify test data based on the  reconstruction error over each dictionary. In \cite{mairal_discriminative_2008}, dictionaries that are both reconstructive and discriminative are learnt for each class by optimizing a sparse reconstruction error term and a discriminative term. The discriminative term in the objective function involves the reconstruction errors of samples over the dictionaries. Test samples are then classified by searching for the dictionary giving the minimum reconstruction error. A smoothing graph cut step is finally applied to refine the label image.
%
%They are simultaneously learned by a method optimizing jointly a sparse reconstruction criteria and a discriminative criteria between classes. The goal is to learn efficient dictionaries to represent the data from their own class but inefficient for the data from the other classes. A parameter controls the compromise between reconstruction and discrimination. . 
%
A dictionary is learnt for each class in \cite{ramirez_classification_2010} with an incoherence criterion imposed on the dictionaries to make them independent. This incoherence term is also used in \cite{kong_dictionary_2012} where an additional dictionary is also learnt in order to capture the patterns common to different classes.

%The objective function is thus composed of a term measuring the reconstruction error of the data on the global dictionary (composed of the concatenation of all the dictionaries) with a sparsity constraint on the coefficients, a term measuring the reconstruction error on the corresponding subdictionary, a term forcing the coefficients on the other subdictionaries to be small, and the incoherence term applied on the dictionaries.
% A global and a local classification are proposed, the decomposition being respectively realized on the global dictionary or on each sub-dictionary. The reconstruction errors, computed by considering each sub-dictionary (plus the dictionary containing the common patterns), are then used to classify the test data by choosing the class with the lowest one.
%In \cite{mairal_discriminative_2008}, dictionaries that are both reconstructive and discriminative are learnt for each class by optimizing a sparse reconstruction criterion and a discriminative criterion between classes. %The goal is to learn efficient dictionaries to represent the data from their own class but inefficient for the data from the other classes. A parameter controls the compromise between reconstruction and discrimination. The discriminative term is applied on the reconstruction errors computed on the dictionaries to make them discriminant for classification.  The classification of a test vector can then be made by searching for the dictionary giving the minimum reconstruction error. \rev{CLAUDE: PARAGRAPHE TRES LONG}

Finally, there are also discriminative dictionary learning methods relying on a categorical or relational organization of the image classes. 
% Cao et al.
A dictionary learning method for multilabel image annotation is proposed in \cite{cao_sled_2015}, where the image labels are first organized into exclusive groups such that two labels that simultaneously occur in the same training image are in different groups. A discriminative dictionary is then learnt with a Fisher criterion for each label group. Test images are finally classified according to their sparse representations by imposing group sparsity in their sparse coding.
The method in \cite{shen_multilevel_2015} learns discriminative dictionaries with a multilevel structure. Their method addresses the particular application of large scale classification with a high number of classes and relies strictly on the availability of a category hierarchy organization of the given classes in the form of a tree model. A global tree-structured dictionary is then learnt where the multilevel tree structure is directly inherited from the given category hierarchy tree model, and the dictionary in each node of the tree is specialized for a group of classes residing under the same subcategory. A similar tree structure is used for emotion classification in \cite{chen_sparse_2015}, where each node is associated with a dictionary and a classifier. While the dictionaries are learnt in an unsupervised manner, the classifiers are trained so as to discriminate between the confused classes branching from that node based on sparse codes. Although the methods in \cite{shen_multilevel_2015} and \cite{chen_sparse_2015} learn tree-structured multilevel dictionaries, these methods differ significantly from ours in that their multilevel dictionary structures are formed quite differently for different usages and purposes.

\section{Learning discriminative structured dictionaries for classification}
\label{sec:proposed_dl_method}

%why structured dicos ?
%why discriminative dicos ?

Our classification method is based on the learning of discriminative structured dictionaries. Multilevel structured dictionaries, composed of many small dictionaries organized on several levels, have the ability to better specialize and thus more efficiently capture the high variability within a class. %We first present the Adaptive Structure, which is a multilevel dictionary structure derived from a tree. We then propose our supervised dictionary learning algorithm, which adopts the Adaptive Structure for learning a discriminative dictionary for each class.

%Besides, It has been shown, for example in \cite{mairal_discriminative_2008}, that using discriminative dictionaries offers better classification results than purely reconstructive ones. This is why we present a method to discriminate our dictionary structures with a new objective function, composed of a reconstructive and a discriminative term, to learn the dictionaries.

%\IEEEpubidadjcol
%\subsubsection{Subsubsection Heading Here}

% ITD (BITD & TSITD)
% Papers ICASSP (Tree K-SVD) & PCS (Adaptive Structure) 2013

This concept of structured dictionaries has been first introduced in \cite{zepeda_iteration-tuned_2010}, and developed in \cite{zepeda_image_2011}, under the name of Iteration-Tuned Dictionaries (ITD). The structure is based on the idea of learning a different dictionary for each iteration of the pursuit algorithm. Thus, each atom added in the decomposition of a signal is selected in a new dictionary by going down the multilevel dictionary strucuture. Several structures, represented in several levels that contain one or several dictionaries, have been developed within this concept, like the Basic ITD (BITD) composed of one dictionary per level, or the Tree-Structured ITD (TSITD) structured as a tree of dictionaries. In these structures, each dictionary at a level is learnt based on a subset of residuals computed at the previous level.
\begin{figure}[!t]
\centering
\includegraphics[trim=1cm 11cm 1cm 1.8cm, width=1\linewidth]{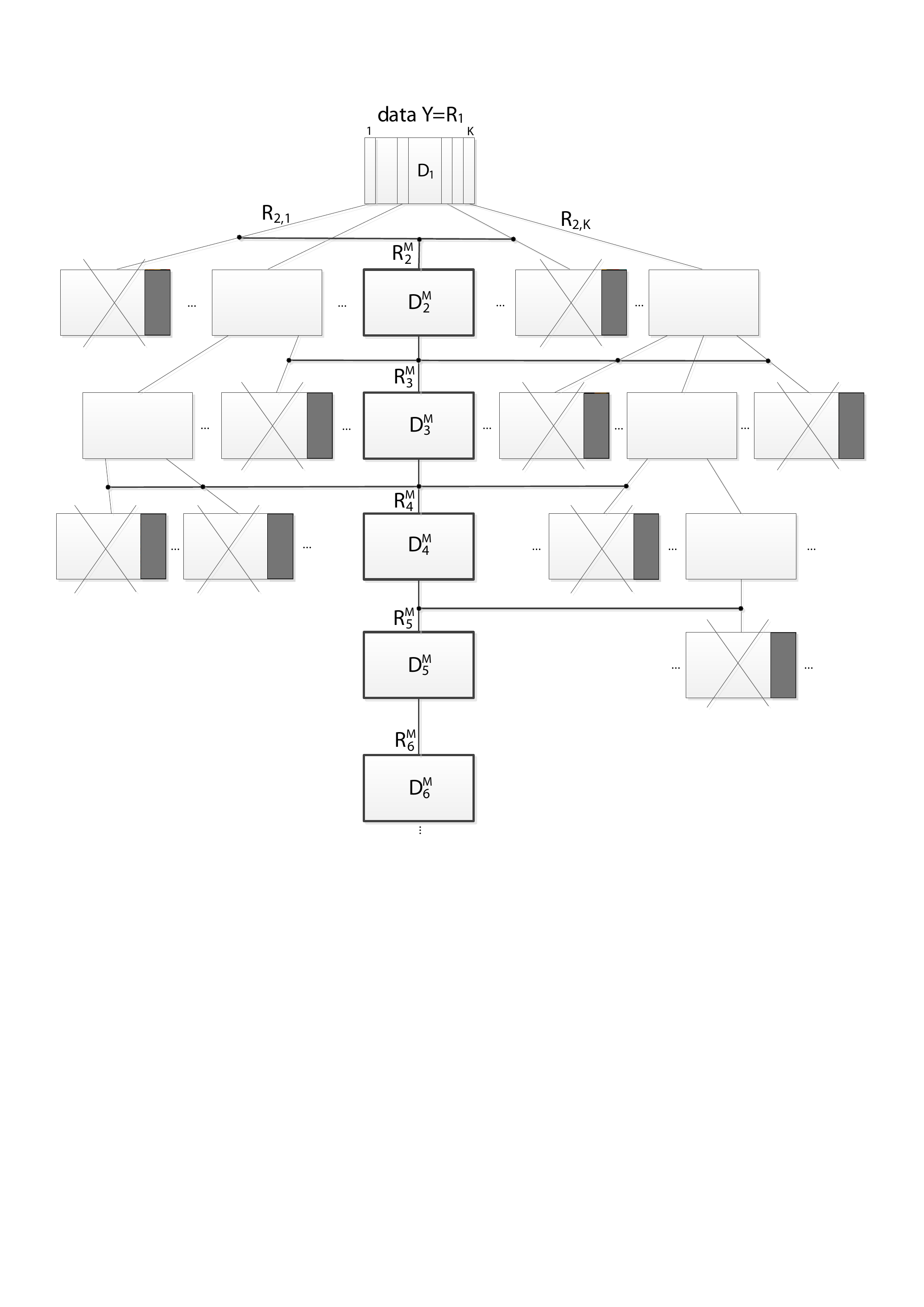}
\caption{The Adaptive Structure. Each atom at a given level leads to the generation of a specialized dictionary at the next level, learnt from only the data samples selecting that atom. At each level $k$, all branches without sufficiently many data samples to continue learning a dictionary at the next level are merged together to learn a new dictionary $\Dm_{k+1}$ at the next level $k+1$.}
\label{fig_AdStr}
\end{figure}
Another tree structure, called Tree K-SVD \cite{aghaei_mazaheri_learning_2013}, has been derived from the TSITD structure. Each dictionary it contains is learnt with the K-SVD algorithm \cite{aharon_k-svd:_2006} with a sparsity of one atom. Starting with one dictionary at the first level, the principle of these tree structures is to learn for each atom at a level one child dictionary at the next level. They are thus quickly composed of too many dictionaries when the number of levels increases, and many can be incomplete or even empty, which can be problematic.

Motivated by these observations, we propose here the discriminative Adaptive Structure by building on our previous work \cite{aghaei_mazaheri_learning_2013-1}, which focused on image compression by learning reconstructive Adaptive Structures. The Adaptive Structure is a new dictionary structure whose topology is adaptively determined during the learning in order to not contain any incomplete dictionary. The branches in the structure are progressively pruned, according to their usage rate, and merged into a unique and more general branch whenever there is not enough data to learn new dictionaries down the branches. This adaptive structure enables the learning of more levels than the tree structure while keeping the total number of atoms reasonable.   In the sequel, we first describe the Adaptive Structure in Section \ref{ssec:adaptive_structure}. Then, in Section \ref{ssec:discrim_structures} we present the proposed discriminative dictionary learning method based on Adaptive Structures in a supervised learning setting. Finally, in Section \ref{sec:sup_class_dl}, we present our supervised classification algorithm.

\subsection{The Adaptive Structure}
\label{ssec:adaptive_structure}

The Adaptive Structure demonstrated in Fig.~\ref{fig_AdStr} is learnt with a top-down approach level after level. Each dictionary in the structure consists of $K$ atoms and is learnt with K-SVD \cite{aharon_k-svd:_2006}, \cite{aharon_k-svd_????}, for a sparsity of one atom. 
%Using the K-SVD algorithm with a sparsity of one atom implies that the Singular Value Decomposition (SVD), usually applied to an error matrix to update each atom, is here directly applied to the training data of the dictionary, with a restriction to the ones using this atom in their decomposition.
Let $Y=\{ y^i\}_{i=1}^N$ denote the set of training samples. The single dictionary at the first level 
\[
D_1=[d_1^1 \ d_1^2 \ \dots \ d_1^K]
\] 
consisting of $K$ unit-norm atoms is learnt using all training data $Y$, by setting the residual term of the first level simply as $R_1=Y$. Each training data sample $y^i$, $i=1, \dots, N$, is then approximated by one atom of the first dictionary $D_1$ as
\[
y^i \approx \langle y^i, d_1^{k_i}  \rangle \, d_1^{k_i},  \text{ with } k_i = \arg \max_j | \langle y^i , d_1^{j} \rangle |
\]
 and the residual vectors
 \[
 r^i = y^i - \langle y^i, d_1^{k_i}  \rangle \, d_1^{k_i} 
 \]
 are computed to form the residual set $R_2$ for the next level. The residuals in $R_2$ are split into $K$ groups $\{ R_{2,k} \}_ {k=1}^K$ such that each group $R_{2,k}$ consists of the residuals of the training samples selecting the atom $d_1^k$ in the first level
\[
R_{2,k} = \{  r^i :   k= \arg \max_j | \langle y^i , d_1^{j} \rangle | \}.
\]
For each set $R_{2,k}$ with $k =1,...,K$, if it contains sufficiently many residuals to satisfy
\[
| R_{2,k} | \geq K
\]
a dictionary at the second level is learnt from $R_{2,k}$, where $| \cdot |$ denotes the cardinality of a set. Otherwise, in order not to create an incomplete dictionary, the dictionary is not learnt and the set of residuals $R_{2,k}$ is saved. At the end of the learning of the second level, all the saved residual sets at this level are merged in $\Rm_2$ as
\[
\Rm_2 = \bigcup_{k:  \ | R_{2,k} | < K}  R_{2,k} .
\]
The merged residual set $\Rm_2$ is then used to learn a new dictionary $\Dm_2$, the  dictionary of the ``merged branches" at the second level of the structure.

The same procedure is then applied to the dictionaries of the second level to learn the dictionaries of the third level. The residual sets of insufficient cardinality at the third level are merged, together with the residuals from $\Dm_2$ at the previous level, to form $\Rm_3$. The residual set $\Rm_3$ is then used to learn the corresponding dictionary $\Dm_3$ at the third level.

This procedure is continued to learn the multilevel Adaptive Structure until a desired number of levels is reached. With this method, the branches with a high usage rate, i.e., the branches selected by many training samples, will be further developed to result in new dictionaries down the tree. On the other hand, the branches with a low usage rate will be quickly pruned and the corresponding residuals will be merged to learn rather general dictionaries (in contrast to the more specialized ones residing at non-merged branches). Thus, during this learning process, the structure adapts itself according to the training vectors in order not to contain any incomplete or empty dictionaries.

Once the Adaptive Structure is learnt, the sparse decomposition of a test sample is computed by selecting one atom per level, beginning with the first level and descending down the multilevel structure. Given a test sample $y$, it is approximated at the first level by an atom $d_1$ of the first-level dictionary selected with the MP algorithm \cite{mallat_matching_1993} with a sparsity of one 
\[
y \approx x_1 d_1
\]
where $x_1$ is the sparse coefficient obtained as
\[
x_1 = \langle y, d_1  \rangle .
\]

The residual vector $y- x_1 d_1$ is then computed and approximated with the same procedure by another atom $d_2$ from a dictionary at the second level, the child dictionary of the atom $d_1$ chosen at the first level. The residual computation and atom selection procedure is continued by descending down the multilevel structure along a branch until a given sparsity is reached. The dictionary to use at each level $l$ is thus determined by the atom $d_{l-1}$ chosen at the previous level in the approximation of $y$. When the end of a branch is reached, the atom  at the next level is selected within the dictionary of the ``merged branch" of the structure, and the decomposition continues after that along this branch. For a structured multilevel dictionary $D$, the reconstruction error of the test sample $y$ for a sparsity of $L$ atoms is thus obtained as
\begin{equation}
\Rerr(y,D) = ||y -  x_1 d_1 - ... -  x_L d_L ||_2^2 
\end{equation}
where $d_1,...,d_L$ are the atoms chosen at the levels $1$ to $L$ from $D$ and $x_1,...,x_L$ are the corresponding coefficients. 

%Besides, the adaptive structure allows the learning of a deeper structure with more levels than in a tree structure, while keeping the dictionary size reasonable, because the branches are progressively pruned. That way, there is no over-fitting. Moreover, the branches are pruned just before reaching an incomplete dictionary, so all the dictionaries in the structure are complete and the deep levels stay efficient, contrarily to the tree structure.

\subsection{Discriminative Learning with Adaptive Structures}
\label{ssec:discrim_structures}

We now describe our proposed method where discriminative Adaptive Structures are learnt for supervised image classification. We propose to learn one multilevel dictionary with the Adaptive Structure for each class. We have observed that in order to achieve satisfactory performance, it suffices to apply the discriminative learning procedure described below at the first level of the structures where there is only one dictionary. We learn the dictionaries at the other levels with the K-SVD algorithm with a sparsity of 1 atom, by following the Adaptive Structure as described in Section \ref{ssec:adaptive_structure}. Since the dictionary structure is learnt with a top-down approach, applying a discrimination-based learning at the top level impacts the other levels as well and has an effect on the whole multilevel dictionary structure.

Let $D^c_1$ denote the dictionary at the first level of the Adaptive Structure to be learnt for the class $c$, for $c = 1,...,C$. We aim to learn a dictionary that is both reconstructive and discriminative, which efficiently represents the data from its own class but yields a large reconstruction error for the data from other classes. Hence, the dictionaries are learnt considering the data from both their own class and the other classes. In this way, the reconstruction errors of test samples on the learnt class-representative dictionaries can be used to classify test data.

In the following, we first introduce our discriminative dictionary learning objective and then discuss its minimization. Next, we explain how the data samples included in the objective function are chosen and finally present the overall discriminative dictionary learning algorithm. 

\subsubsection{Discrimination model} 

We propose to update the dictionaries sequentially (atom by atom), by minimizing the following objective function for updating an atom $d \in \mathbb{R}^n$ of the dictionary $D^c_1$ of class $c$

\begin{equation}
\underset{d \in \mathbb{R}^n}{\min} [ ||Y_c^R - d d^T Y_c^R||_F^2 - \lambda ||Y_{c_{\neq}} - d d^T  Y_{c_{\neq}}||_F^2 ] \; \textrm{with} \; ||d||_2 = 1.
\label{eq:pb_optim_classif}
\end{equation}
The first term in the above cost function is a reconstructive term aiming to adapt the atom to the training data $Y_c$ from its own class $c$, where  $Y_c^R$ denotes the restricted subset of data samples from class $c$ that use the atom $d$ in their decomposition. The second term is a discriminative term, whose goal is to push the atom $d$ away from the training samples $Y_{c_{\neq}}$ of the other classes. Thus, we search for the atom $d$ minimizing the reconstruction error of the data from its own class and maximizing the reconstruction error of the data from the other classes.
The positive weight parameter $\lambda$ balances the two terms according to the ratio between the number of samples in $Y_c^R$ and $Y_{c_{\neq}}$ as 
\begin{equation}
\lambda = \frac{ | Y_c^R |}{ | Y_{c_{\neq}} |}  \ \alpha
\end{equation}
where $| \cdot |$ denotes the number of columns in a matrix (i.e., the number of data samples) with a slight abuse of notation. The positive constant $\alpha$ adjusts the compromise between reconstruction and discrimination.  The exact choice of the samples $Y_{c_{\neq}}$ for each class $c$ and atom $d$ will be explained later in Section \ref{sssec:aff_matrix}.

\subsubsection{Minimization of the objective function} 

The cost function to minimize can be rewritten as
\begin{multline}
\underset{d \in \mathbb{R}^n}{\min} [ \tr( (Y_c^R - d d^T Y_c^R)^T (Y_c^R - d d^T Y_c^R) ) \\ 
- \lambda \tr( (Y_{c_{\neq}} - d d^T  Y_{c_{\neq}})^T (Y_{c_{\neq}} - d d^T  Y_{c_{\neq}}) )].
\end{multline}
This is equivalent to
\begin{multline}
\underset{d \in \mathbb{R}^n}{\min} [ \tr({Y_c^R}^T Y_c^R - 2 {Y_c^R}^T d d^T Y_c^R  
+ {Y_c^R}^T d d^T d d^T Y_c^R) \\
- \lambda \tr(Y_{c_{\neq}}^T Y_{c_{\neq}} - 2 Y_{c_{\neq}}^T d d^T Y_{c_{\neq}} + Y_{c_{\neq}}^T d d^T d d^T Y_{c_{\neq}}) ].
\end{multline}
With the constraint $||d||_2 = 1$, we have $d^T d = 1$. Hence, we can simplify the cost function to
\begin{equation}
\underset{d \in \mathbb{R}^n}{\min} [ \tr({Y_c^R}^T Y_c^R - {Y_c^R}^T d d^T Y_c^R )
- \lambda \tr(Y_{c_{\neq}}^T Y_{c_{\neq}} - Y_{c_{\neq}}^T d d^T Y_{c_{\neq}}) ].
\end{equation}
In order to solve this minimization problem under the constraint $d^Td=1$, we then apply the Lagrange multipliers method and minimize the function $L(d,\mu)$
\begin{multline}
L(d,\mu) = \tr({Y_c^R}^T Y_c^R - {Y_c^R}^T d d^T Y_c^R) \\
- \lambda \tr(Y_{c_{\neq}}^T Y_{c_{\neq}} - Y_{c_{\neq}}^T d d^T Y_{c_{\neq}})
+ \mu (d^T d - 1).
\end{multline}
Setting the derivative of $L(d,\mu)$ with respect to $\mu$ to $0$ gives
\begin{equation}
 d^T d = 1.
\end{equation}
We then evaluate the derivative with respect to $d$ and equate it to $0$ as
%\begin{equation}
%\frac{\partial L}{\partial d} = 0.
%\end{equation}
%The derivative with respect to $d$ is evaluated as
\begin{equation}
\begin{split}
\frac{\partial L}{\partial d} 
&= \frac{\partial}{\partial d} \tr(- {Y_c^R}^T d d^T Y_c^R)  
- \lambda \frac{\partial}{\partial d} \tr(- Y_{c_{\neq}}^T d d^T Y_{c_{\neq}})
+ 2 \mu d^T   \\
 &= -2 d^T Y_c^R {Y_c^R}^T
+ 2 \lambda d^T Y_{c_{\neq}} Y_{c_{\neq}}^T
+ 2 \mu d^T  \\
&= d^T (-2 Y_c^R {Y_c^R}^T + 2 \lambda Y_{c_{\neq}} Y_{c_{\neq}}^T + 2 \mu I)  =  0
\end{split}
\end{equation}
which gives
\begin{equation}
d^T (\lambda Y_{c_{\neq}} Y_{c_{\neq}}^T - Y_c^R {Y_c^R}^T) = - \mu d^T.
\end{equation}
Taking the transpose of both sides, we get
\begin{equation}
(Y_c^R {Y_c^R}^T - \lambda Y_{c_{\neq}} Y_{c_{\neq}}^T) d = \mu d.
\end{equation}
This equation is of the form
\begin{equation}
Ad = \mu d
\end{equation}
with
\begin{equation}
\label{eq:obj_A_matrix}
A = Y_c^R {Y_c^R}^T - \lambda Y_{c_{\neq}} Y_{c_{\neq}}^T.
\end{equation}

The atom $d$ is thus an eigenvector of $A$ with a unit $\ell_2$-norm. Since our objective in \eqref{eq:pb_optim_classif} imposes the atom to fit the samples $Y_c^R$ while repulsing it from $Y_{c_{\neq}}$, the sought atom is the eigenvector of $A$ corresponding to its maximum eigenvalue.

 %Notice that by choosing $\lambda = 0$, we obtain back the K-SVD update, which computes the SVD of the data matrix $Y^R$ (for a sparsity of one atom) and chooses the first left singular vector, i.e., the eigenvector of $Y^R {Y^R}^T$ corresponding to the maximum eigenvalue.
%We apply here the same procedure and update $d$ with the eigenvector of $A$ corresponding to the maximum eigenvalue.

\subsubsection{Choice of the sample set $Y_{c_{\neq}}$} 
\label{sssec:aff_matrix}

In order to adapt the discrimination term to each class and even to each atom to update, we follow a particular strategy when forming the matrix $Y_{c_{\neq}}$ that contains the data from the other classes than the current class $c$. Rather than choosing $Y_{c_{\neq}}$ to contain all data samples from the other classes, we wish to particularly discriminate the updated atom from the classes most similar to its class. For this purpose, we compute an affinity matrix that represents the similarity between each pair of classes.

In order to compute the class affinity matrix, for each class $c$ we first compute a representative vector $\eta_c$ that best fits the data samples $Y_c$. In order to avoid computing an almost constant vector as the representative vector, we first subtract the mean value of each data sample $y_c^i$ in $Y_c$. We then choose the representative vector as the one that maximizes the energy of the data samples when projected onto it as
\[ 
\eta_c = \arg \max_\eta \sum_i ( \langle \overline y_c^i, \eta  \rangle  )^2 \text{ s.t. } \| \eta \|_2 = 1
\]
where $ \overline y_c^i $ is the mean-removed version of the training sample $y_c^i$. The solution of the above problem gives $\eta_c$ as the unit-norm eigenvector of $\overline Y_c \overline Y_c^T$ associated with its maximum eigenvalue, where $\overline Y_c$ is the matrix containing the mean-removed samples $\{ \overline y_c^i \}$. We then obtain the class affinity matrix $S \in \mathbb{R}^{C \times C}$ such that the affinity $S_{ij}$ between the $i$-th and $j$-th classes is given by the similarity between their class representative vectors as
\[
S_{ij} = |  \langle \eta_i , \eta_j \rangle |.
\]
%
%The absolute value of the inner product between each representative vector is then computed to get for each class an affinity measure with each other class. These values are stored in an affinity matrix $AM$ of size $C \times C$ (with $C$ the number of classes). 
Hence, $S$ is a symmetric matrix with $1$'s on the diagonal and affinity values varying between $0$ and $1$ on its off-diagonal entries.

With this affinity matrix, we then determine $Y_{c_{\neq}}$ by selecting from each class $j$ a variable number of vectors according to its affinity $S_{jc}$ with the current class $c$. If the number of training samples in each class is the same and equal to $N$, we set the number of samples to be selected from class $j$ as
\[
| Y^j_{c_{\neq}} | = round(  S_{jc} \, N )
\]
where the $round$ function rounds the values to the nearest integer. Note that this strategy can be easily adapted to the case where the number of training samples is different for each class, by choosing $ | Y^j_{c_{\neq}} | $ such that
%We determine the number of vectors to be selected from each class such that
%
\[
\frac{ | Y^j_{c_{\neq}} |}{ | Y_{c_{\neq}} |} 
\approx
\frac{S_{jc} }{\sum_{j} S_{jc}}
\]
where $Y^j_{c_{\neq}} $ contains the samples in $Y_{c_{\neq}}$ from class $j$, with $j \neq c$. The samples $Y^j_{c_{\neq}}$ are chosen as the samples from class $j$ that have the highest correlation with the atom $d$ to update, i.e., the samples from class $j$ that are the most susceptible to choose $d$ for their sparse decomposition in the dictionary $D^c$ of class $c$.

% fact, the values in the affinity matrix, multiplied by $100$, give the percentages of training vectors selected in each class other than the current class $c$, and concatenated in $Y_{c_{\neq}}^{R_{AM}}$. For each class, the most correlated training vectors to the atom $d$ to update are chosen. 

% apply a restriction $R_{AM}$ on the vectors from $Y_{c_{\neq}}$. Instead of selecting all the training vectors from the other classes ($Y_{c_{\neq}}$), we select for each class a variable number of vectors according to its affinity with the current class $c$, measured in the affinity matrix, in order to create the matrix $Y_{c_{\neq}}^{R_{AM}}$. In fact, the values in the affinity matrix, multiplied by $100$, give the percentages of training vectors selected in each class other than the current class $c$, and concatenated in $Y_{c_{\neq}}^{R_{AM}}$. For each class, the most correlated training vectors to the atom $d$ to update are chosen.

With this strategy, each dictionary becomes more discriminative towards the classes closest to its own class, instead of equally treating all the other classes. Indeed, two classes with high dissimilarity do not necessarily need an extra discrimination criterion to be distinguished.

%With this new restriction, the cost function to minimize becomes:
%\begin{multline}
%\underset{d \in \mathbb{R}^n}{\min} [ ||Y_c^R - d d^T Y_c^R||_F^2 - \lambda ||Y_{c_{\neq}}^{R_{AM}} - d d^T  Y_{c_{\neq}}^{R_{AM}}||_F^2 ]  \\
%\; \textrm{with} \; ||d||_2 = 1 
%\end{multline}
%where:
%\begin{equation}
%\lambda = \frac{\#Y_c^R}{\#Y_{c_{\neq}}^{R_{AM}}} \times \alpha
%\end{equation}
%and $R_{AM}$ is the restriction on $Y_{c_{\neq}}$ due to the affinity matrix.
%The matrix $A$ becomes then:
%\begin{equation}
%A = Y_c^R {Y_c^R}^T - \lambda Y_{c_{\neq}}^{R_{AM}} {Y_{c_{\neq}}^{R_{AM}}}^T
%\end{equation}

\subsubsection{Overall discriminative dictionary learning algorithm} %to learn a discriminative dictionary
 
Let us now describe the overall algorithm to learn a multilevel discriminative dictionary for each class $c$.

The dictionary $D^c_1$ that composes the first level of the Adaptive Structure for class $c$ is computed as follows. The dictionary is first initialized by training vectors from its own class, randomly selected and normalized to be of unit $\ell_2$-norm. The algorithm iterates between a sparse decomposition step and a dictionary update step as frequently done.

In the sparse decomposition step, the decompositions of the data samples $Y_c$ from class $c$ are computed with the MP algorithm \cite{mallat_matching_1993} for a sparsity of one atom. Thus, for each vector in $Y_c$, we search for the atom in $D^c_1$ that is the most correlated with it. This step will allow us to compute for each atom $d$ the matrix $Y_c^R$ composed of the training vectors from the class $c$ choosing this atom $d$ at this decomposition step.

The dictionary is then updated sequentially, atom by atom. For each atom $d$ of $D^c_1$, the matrix $Y_{c_{\neq}}$ composed of training vectors from the other classes is formed with respect to the class affinities as described in Section \ref{sssec:aff_matrix} and the matrix $Y_c^R$ is computed. The matrix $A$ in \eqref{eq:obj_A_matrix} can then be computed and the atom $d$ is updated as the unit-norm eigenvector of $A$ corresponding to the maximum eigenvalue.

Once the discriminative dictionary $D^c_1$ is computed by alternatingly updating the sparse codes and the atoms, the residual set for the next level is computed from $Y_c$, and the reconstructive Adaptive Structure learning described in Section \ref{ssec:adaptive_structure} continues until the desired number of levels. This procedure is repeated for each class $c$ to obtain a class-representative multilevel structured dictionary $D^c$ for each class.

%residuals are split into $K$ groups. For each residual group, a new dictionary is learned at the second level and the residuals are merged when necessary with respect to the Adaptive Structure as described in Section \ref{ssec:adaptive_structure}. This procedure is repeated for each class $c$ to obtain class-representative multilevel structured dictionary $D^c$ for each class. 

\subsection{Classification of test images based on learnt structured dictionaries}
\label{sec:sup_class_dl}

%In supervised classification, the labels of the classes are known and some exemplars are available for each class. This supervised classification method is based on the learning of one dictionary per class. As shown in \cite{mairal_supervised_2008}, adding a discriminative criterion in the learning makes the dictionaries more adapted to classification and improve the results compared to purely reconstructive dictionaries. In this paper, we learn structured dictionaries with a trade-off between reconstruction and discrimination capabilities, as we will explain in Section III.

Test samples are classified with respect to their reconstruction errors over the learnt multilevel dictionaries as follows. A given test sample $y$ is first decomposed over each one of the $C$ class-representative dictionaries for a given sparsity $L$, where $C$ is the number of classes. Then the reconstruction error of $y$ is computed over each dictionary $D^c$ of class $c$, $c =1,...,C$, as described in Section \ref{ssec:adaptive_structure}
\begin{equation}
\label{eq:R}
R(y,D^c) = ||y - x_1^c d_1^c - \dots - x_L^c d_L^c||_2^2.
\end{equation}
Here $d_l^c $ is the atom selected at level $l$, chosen in the dictionary at level $l$ that corresponds to the atom $d_{l-1}^c$ selected at the previous level $l-1$; and  $x_l^c$ is the coefficient of $d_l^c$ in the decomposition of $y$.

 %as explained in Section \ref{}, by choosing the atom at each level $l$ within the dictionary at level $l$ that corresponds to the previous atom  then selecting the atom   $x$ the coefficient vector containing $L$ non-zero values that is computed by selecting an atom at each level of
%The decomposition on a structured dictionary is a bit different as one atom is selected per level, with the MP algorithm. 

In this paper, we focus on pixelwise classification of a test image. In this case, the training and test samples are image patches. The test samples $y$ are obtained by taking a square patch around each pixel of the given test image so as to assign a class label to each pixel. In such a setting, it is useful to normalize the reconstruction residuals by the norm of the test patch $y$, in order to prevent the patches of high norm from dominating the overall label estimation during the smoothing steps discussed in Section \ref{sec:smoothing}. We thus consider the normalized error 
\begin{equation}
\label{eq:norm_rec_err}
E(y,D^c) = \frac{R(y,D^c)}{||y||_2^2}.
\end{equation}
A simple classification strategy would be to search the class $\hat{c}$ minimizing the error $E(y,D^c)$  for each patch $y$
\begin{equation}
\label{eq:class_rule_rec_err}
\hat{c} = \underset{c=1,...,C}{\arg \min} \; E(y,D^c).
\end{equation}
However, this classification rule leads in general to a fractional segmentation of the test picture resulting in many small and disconnected label support regions. In order to improve the label image and obtain more uniform and smooth label supports, we apply two smoothing steps, discussed in Section \ref{sec:smoothing}.

%, from the error values $Err(y,D_c)$, to regroup the small areas together and thus clean the classification. The chosen smoothing methods are presented in Section IV.

\section{Smoothing steps}
\label{sec:smoothing}

%A simple classification method would be to select for each pixel the class whose dictionary offers the minimum reconstruction error for the patch considered around the pixel. But this kind of classification leads to a noisy segmentation with many small labels. That is why it is necessary to apply a smoothing method to obtain a clean segmentation of the test picture in a few big labels.
In order to improve the estimate of the label image obtained via the reconstruction errors over the class-representative dictoinaries as described in \eqref{eq:class_rule_rec_err}, we apply a label smoothing procedure that comprises two steps: a label expansion step with a graph cut, followed by an erosion step to erode the remaining small undesirable label support regions.

\subsection{Label expansion via graph cuts}

The first smoothing step considers an $\alpha$-expansion algorithm minimizing an energy function with a graph cut \cite{boykov_fast_2001, kolmogorov_what_2004, boykov_experimental_2004}.
The algorithm estimates the label $f_p$ of each pixel $p$ by minimizing the following energy function based on a Potts model:
\begin{equation}
J(f) = \sum_{p \in P} C_p (f_p) + \sum_{ p \sim q; \ p, q \in P} \upq \, T(f_p, f_q).
\label{energy_fct}
\end{equation}

The first term is the data cost and corresponds to the sum on all the pixels $P$ of the cost $C_p(f_p)$ of assigning a label $f_p$ to the pixel $p \in P$. 
%These costs are given by a matrix $Data_{cost}$ containing, for each pixel of the picture to classify, a cost associated to each class (or label). 
Rather than applying a cost of $0$ to the class offering the lowest reconstruction error, and $1$ to the other classes as done in \cite{mairal_discriminative_2008}, we set the data cost as
\[
C_p (f_p) = E(y_p, D^{f_p})
\]
where $y_p$ is the patch centered around the pixel $p$ and $E(y_p, D^{f_p})$ is the reconstruction error of $y_p$ over the dictionary of the class $f_p$ as defined in \eqref{eq:norm_rec_err}. Such a choice of the data cost provides a ranking of all the classes for each pixel. 
%  in $Data_{cost}$ for each pixel $p$ the values $Err(y,D_c)$ (one value per class $c$ for each patch $y$ around a pixel $p$). That way, all the classes are ranked for each pixel. 

The second term is the smoothing cost summed over all neighboring pixels $p$ and $q$ (denoted as $p \sim q$), which is defined as
\begin{align*}
T(f_p, f_q)=\begin{cases}
1 &\text{ if } f_p \neq f_q \\
0 &\text{ if } f_p = f_q.
\end{cases} 
\end{align*}
%
%$T(f_p \neq f_q)$ equals $1$ if the label $f_p$ is different from the label $f_q$ and $0$ otherwise. 
Hence, if two neighboring pixels $p$ and $q$ share the same label, then the associated cost is $0$. Otherwise, this cost is constant and equal to the parameter $\upq$.

This model encourages a labeling with several large regions whose pixels share the same label. Adapting the parameter $\upq$ makes the label image more smooth or less smooth. By setting it to $0$, only the data cost, i.e. the reconstruction errors, is considered and the resulting label image is composed of many small and disconnected regions as label supports. Meanwhile, choosing a too big $\upq$ will fuse the label supports too much and the estimated label image will contain less label support regions than desired. The parameter $\upq$ can possibly be chosen as a constant or depending on the class labels $f_p$ and $f_q$.
 
An $\alpha$-expansion method \cite{boykov_fast_2001} is applied to minimize the energy function. This method expands at each iteration label after label, searching for the optimal expansion for each label, in order to decrease the energy function $J(f)$. The expansion consists of modifying possibly numerous pixels simultaneously by assigning the current label, called label $\alpha$, to these pixels.
We have used the Matlab wrapper \cite{bagon_matlab_2006} for the experiments.

\begin{figure*}[!htb]
  \centering
  %\addtocounter{subfigure}{4}
  \subfloat[Test image 1]{\includegraphics[trim=0cm 0cm 0cm 0cm,clip=true,scale=0.28]{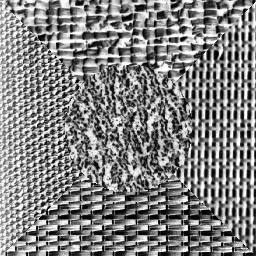}}
  \hskip1em
  \subfloat[Test image 2]{\includegraphics[trim=0cm 0cm 0cm 0cm,clip=true,scale=0.28]{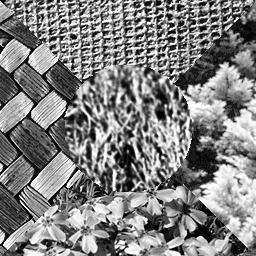}}
  \hskip1em
  \subfloat[Test image 3]{\includegraphics[trim=0cm 0cm 0cm 0cm,clip=true,scale=0.28]{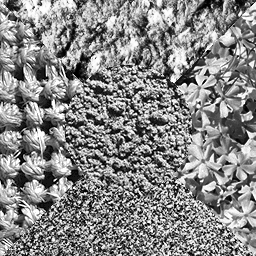}}
  \hskip1em
  \subfloat[Test image 4]{\includegraphics[trim=0cm 0cm 0cm 0cm,clip=true,scale=0.28]{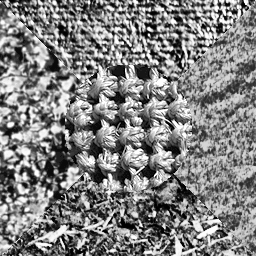}}
  \hskip1em
  \subfloat[Test image 5]{\includegraphics[trim=0cm 0cm 0cm 0cm,clip=true,scale=0.28]{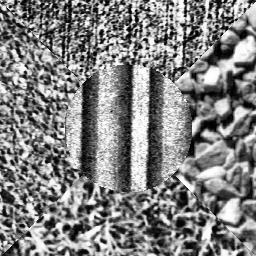}}
  \\
  \subfloat[Test image 6]{\includegraphics[trim=0cm 0cm 0cm 0cm,clip=true,scale=0.28]{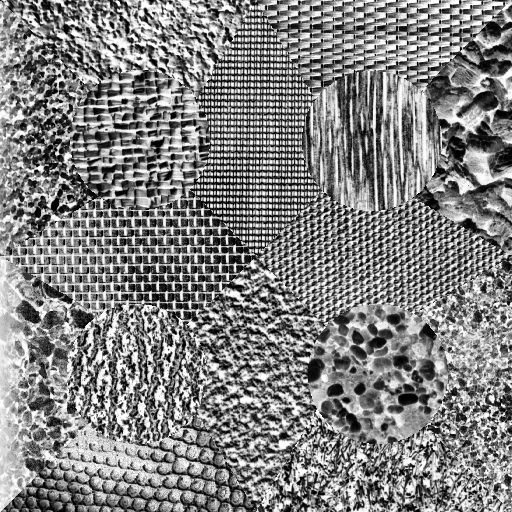}}
    \hskip1em
  \subfloat[Test image 7]{\includegraphics[trim=0cm 0cm 0cm 0cm,clip=true,scale=0.28]{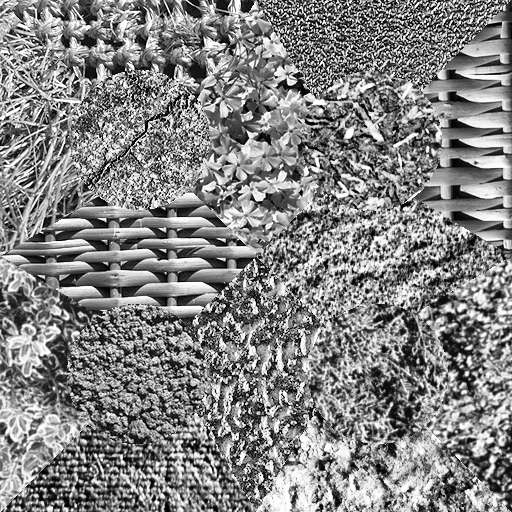}}
  \\
  \subfloat[Test image 8]{\includegraphics[trim=0cm 0cm 0cm 0cm,clip=true,scale=0.28]{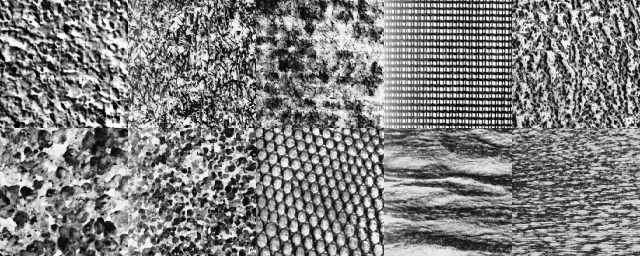}}
  \hskip1em
  \subfloat[Test image 9]{\includegraphics[trim=0cm 0cm 0cm 0cm,clip=true,scale=0.28]{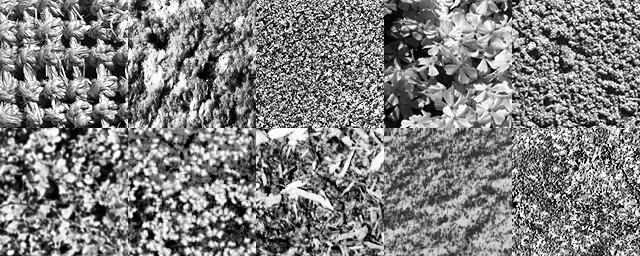}}
  \\
  \subfloat[Test image 10]{\includegraphics[trim=0cm 0cm 0cm 0cm,clip=true,scale=0.28]{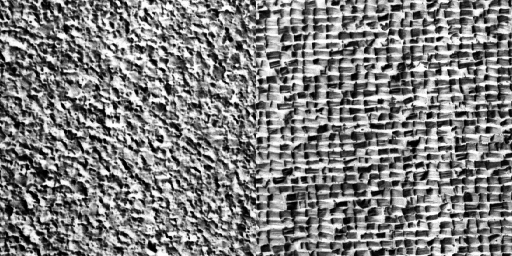}}
  \hskip1em
  \subfloat[Test image 11]{\includegraphics[trim=0cm 0cm 0cm 0cm,clip=true,scale=0.28]{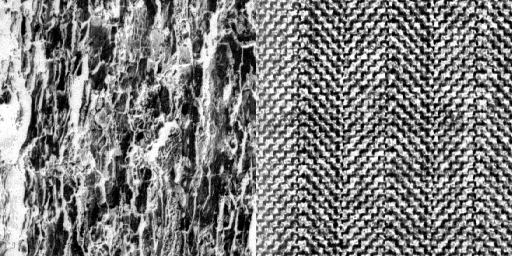}}
 % \\
   \hskip1em
  \subfloat[Test image 12]{\includegraphics[trim=0cm 0cm 0cm 0cm,clip=true,scale=0.28]{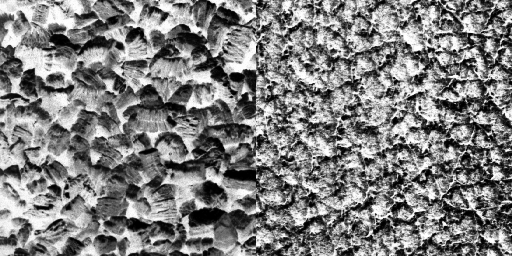}}
  \\
   \vspace{0cm}
  \caption[Texture image dataset: test images]
  {Texture image dataset: test images.}
   \label{fig:Texture_ImagesTest}
\end{figure*}

\subsection{Erosion}

After a first smoothing step realized with a label expansion algorithm, some small undesirable label support regions can remain in the label image. In order to remove them, we add an erosion step \cite{skretting_energy_2014} applied directly on the label image obtained after the first smoothing step and based on the same data cost $C_p (f_p)$. 

The $\alpha$-erosion algorithm \cite{skretting_energy_2014} (available online \cite{skretting_page}) works with a close variant of the energy function \eqref{energy_fct} and seeks to erode the small segments in priority (a segment corresponds to a group of connected pixels of the same label) to decrease the energy function. Too small segments are always eroded whereas too big segments are never eroded. The segments between these limits are treated one after the other, beginning with the smaller ones. For each segment, its pixels are relabeled one by one, the segment being eroded by the segments around. If the new labeling of the segment decreases the energy function, then the erosion of the segment is accepted, otherwise it is canceled.

A slight erosion step is also realized on the label support edges.

\section{Experimental results}
\label{sec:exp_results}

The proposed method is tested on a set of texture images, commonly used for supervised classification and texture segmentation. We compare our method to several state-of-the-art dictionary learning and texture classification algorithms.

\subsection{Pre-processing of the data}

In order to improve the classification, some pre-processing operations, already used in \cite{mairal_discriminative_2008}, are performed on the training and test patches.
A Gaussian mask of standard deviation $4$ is first applied with an element-wise multiplication with the patch, in order to give more weight to the center of the patches, as it is possible that the peripheral pixels of a patch are from a different class if the patch lies on an edge between two classes. The weight  is thus $1$ at the central pixel(s) of the patch and decreases with the distance to the center of the patch. Each patch is then sharpened with a Laplacian filter (of size $3\times3$), each Laplacian filtered patch being subtracted from the patch to get the sharpened patch.
Note that when computing the affinity matrix $S$ as described in Section \ref{sssec:aff_matrix}, we use the non-processed (original) versions of the image patches, as we have observed that pre-processing may impair the estimation of the affinities between classes. 

Besides, in order to be able to classify the pixels on the borders of the test picture, we generate some additional pixels along the borders by taking the mirror image of the pixels close to the borders. This allows the classification of the border pixels by using the square patches artificially generated around them.

\subsection{Texture classification}

\subsubsection{Texture image dataset}

The dataset composed of texture images, used in our experiments, has first been introduced in \cite{randen_filtering_1999}, and has since been used in several articles dealing with classification and segmentation. It has been created with pictures from the Brodatz album \cite{brodatz_textures:_1966}, from the \textit{Vision Texture} database of the MIT, and from the texture image database \textit{MeasTex}. The pictures have thus been captured with different equipments under different conditions.
Each one of the 12 test images in Fig. \ref{fig:Texture_ImagesTest} corresponds to a different supervised classification problem with different texture classes. The number of classes in each problem varies between $2$ and $16$. The training images corresponding to each one of these 12 supervised classification problems are also available. 
%It is composed of $12$ test images (Fig. \ref{fig:Texture_ImagesTest}), composed of $2$ to $16$ textures. Each test image corresponds to a different supervised classification problem with its own classes. So for each test image, the training images corresponding to the classes in the test image are available. 
The training and test images have been taken from different portions of each texture. The dataset is available online \cite{randen_trygve_????}.
%note = {\textit{Available at: http://www.ux.uis.no/$\sim$tranden/}}

\subsubsection{Parameters}

For each one of the 12 texture classification problems, an Adaptive Structure is learnt per texture class from the corresponding training picture. Overlapping $8\times8$ blocks are extracted from these $256\times256$ pictures to learn each dictionary structure on $62001$ training vectors. The structures are composed of complete dictionaries of $64$ atoms. We limit the size of each dictionary in the structures so that they capture the characteristics of their own class and do not become too efficient for the representation of other classes.
The first level of the structures, made discriminant, is learnt in $50$ iterations, whereas the next levels are learnt in $10$ iterations. The dictionaries in the structures are initialized with randomly chosen training vectors. The parameter $\alpha$ balancing the reconstruction and discrimination at the first level is empirically set to $\frac{1}{40}$.

The test patches, also of size $8\times8$ pixels, are decomposed over each class-representative dictionary structure of the corresponding texture classification problem, for a sparsity of $2$ atoms. Since in a classification problem we do not look for the best approximation of a patch but rather would like to classify it based on its reconstruction errors over the different dictionary structures, it is better to avoid high sparsity values. In practice, we have observed that the sparsity of $2$ atoms gives good results in general.

For the first smoothing step with the graph cut, the smoothing parameter $\upq$ is experimentally set to the constant value $0.16$ for all different label pairs $f_p$, $f_q$ with $f_p \neq f_q$, and to $0$ for $f_p = f_q$, which has been observed to yield good results. Finally, for the second smoothing step of erosion, the parameter\footnote{This parameter $\lambda$ is used in the energy function in \cite{skretting_energy_2014} and is different from the parameter $\lambda$ we use to balance reconstruction and discrimination in our objective function to learn each dictionary at the first level of the structures.} $\lambda$  is set to $2$ as in \cite{skretting_energy_2014}. Areas of less than $2000$ pixels are always eroded whereas areas of more than $10000$ pixels never are. Between these limits, the erosion depends on the minimization of the cost function. A slight erosion of $2$ pixels is also performed on the edges.

\subsubsection{Results}

\begin{figure}[t]
  \centering
  \subfloat[Class 1]{\includegraphics[scale=0.5]{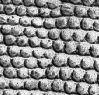}}
  \hskip1em
  \subfloat[Class 12]{\includegraphics[scale=0.6]{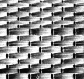}}
  \\
  \subfloat[Multilevel dictionary for class 1]{\includegraphics[scale=0.2]{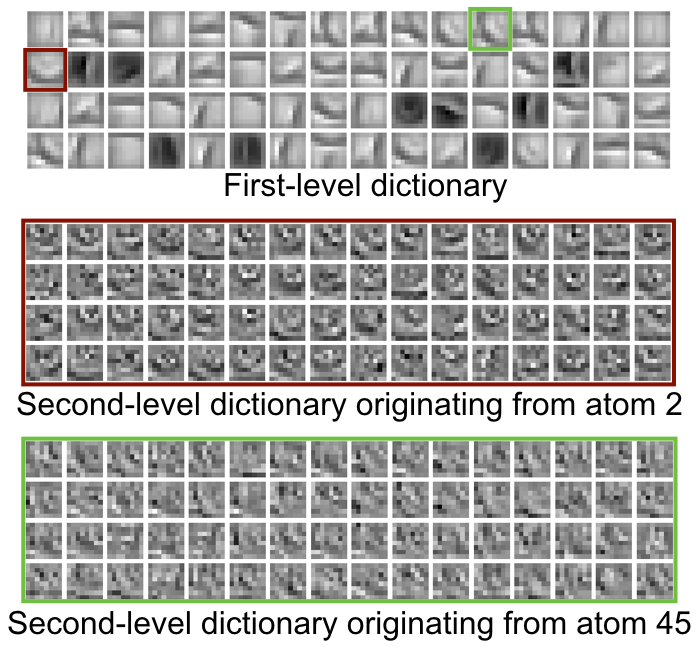}}
  %\hskip1em
  \subfloat[Multilevel dictionary for class 12]{\includegraphics[scale=0.2]{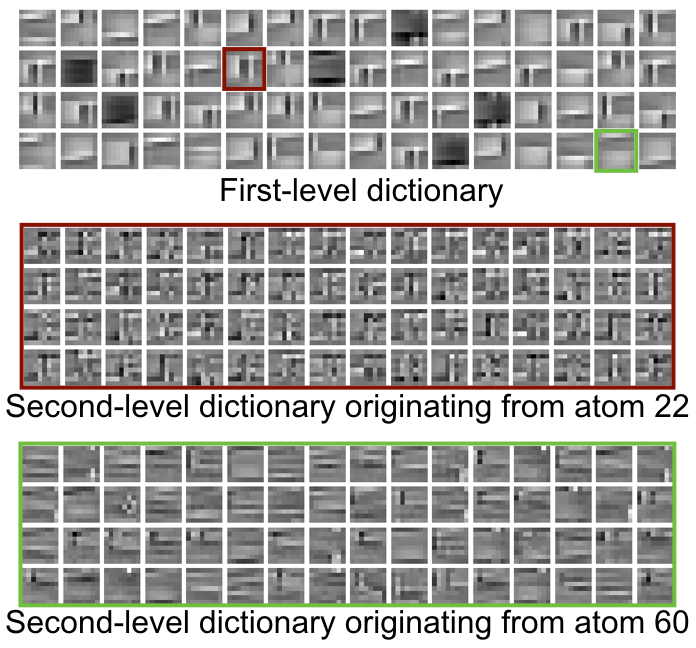}}
  \\
   \vspace{0cm}
  \caption[Multilevel dictionaries learnt for classes 1 and 12 of the experiment 6]
  {Some example multilevel dictionaries learnt for classes 1 and 12 of  experiment 6}
   \label{fig:learnt_multilevel_atoms}
\end{figure}

We first present  in Fig.~\ref{fig:learnt_multilevel_atoms} some example atoms from the multilevel dictionaries learnt with the proposed algorithm for the classification problem of experiment 6 (Fig.~\ref{fig:Texture_ImagesTest}(f)). Sample regions from the training images of the texture classes 1 and 12 of this experiment are shown in Figures \ref{fig:learnt_multilevel_atoms}(a) and \ref{fig:learnt_multilevel_atoms}(b). Figures \ref{fig:learnt_multilevel_atoms}(c) and \ref{fig:learnt_multilevel_atoms}(d) show some of the multilevel dictionaries learnt for these two texture classes. For both texture classes, the first-level dictionary is displayed, together with the second-level dictionaries originating from two different atoms of the first-level dictionary.

It can be observed that the atoms in the first-level dictionaries capture well the main characteristics of each class. The first-level dictionary of class 1 consists of atoms containing rather smooth and curvy features, whereas the  first-level dictionary of class 12 contains atoms capturing straight edges and corners. We can observe that, due to the large intra-class variation in these texture classes rich in content, the atoms in the first-level dictionary of the same class can be quite different from each other. The proposed multi-level dictionary structure is then seen to be well-adapted to this setting as it allows the specialization of the atoms at later levels based on the structure of the atoms at earlier levels they originate from. Indeed, it can be seen in Fig.~\ref{fig:learnt_multilevel_atoms}  that the second-level dictionaries derived from two different first-level atoms of the same class capture finer details but tend to have different characteristics. In class 1, the second-level dictionary derived from atom 2 of the first level inherits the round-shaped circular structure of atom 2, while the second-level dictionary derived from atom 45 is tuned to represent more straight and diagonally-oriented texture features. Similarly, in class 12,  the second-level dictionary originating from atom 60 contains mainly horizontally oriented atoms as the dominant orientation of the atom 60 is horizontal, while the second-level dictionary originating from atom 22 captures both vertically and horizontally oriented corner-like features just like the atom 22. This confirms that the dictionaries learnt at different levels are successfully specialized to adapt to different fine and coarse texture features present in image classes of large intra-class variation.

\begin{figure}[t]
  \centering
  \subfloat[Test image 4]{\includegraphics[trim=1.75cm 1.25cm 1.75cm 0.75cm,clip=true,scale=0.3]{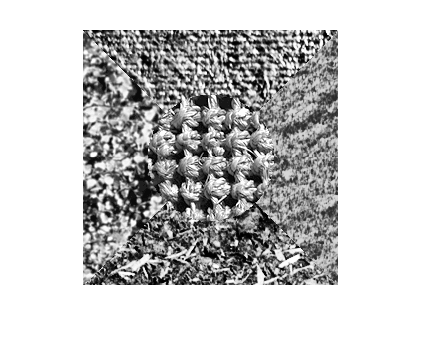}}
  \hskip1em
  \subfloat[Min error]{\includegraphics[trim=1.75cm 1.25cm 1.75cm 0.75cm,clip=true,scale=0.3]{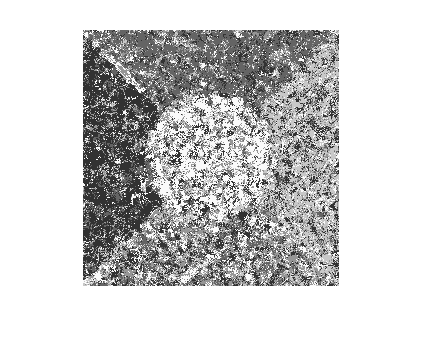}}
  \\
  \subfloat[After graph cut]{\includegraphics[trim=1.75cm 1.25cm 1.75cm 0.75cm,clip=true,scale=0.3]{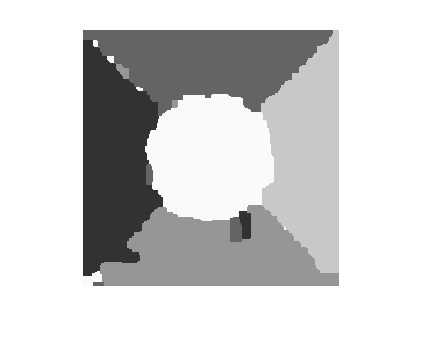}}
  %\hskip1em
  \subfloat[After erosion]{\includegraphics[trim=1.75cm 1.25cm 1.75cm 0.75cm,clip=true,scale=0.3]{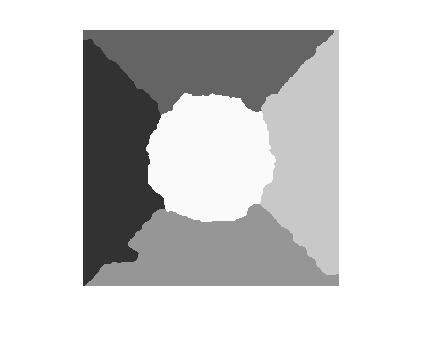}}
  %\hskip1em
  \subfloat[Ground truth]{\includegraphics[trim=1.75cm 1.25cm 1.75cm 0.75cm,clip=true,scale=0.3]{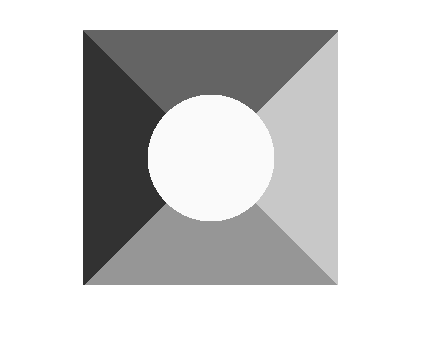}}
  \\
   \vspace{0cm}
  \caption[Test image 4 at the successive steps of the classification algorithm.]
  {Test image 4 at the successive steps of the classification algorithm.}
   \label{fig:Im4_etapes}
\end{figure}

%Example: Image 4
Next, we demonstrate the effect of the different stages of the proposed method. In Fig.~\ref{fig:Im4_etapes} the label images obtained after each step of our classification method are shown for the test image $4$. We can see the benefits of the smoothing steps whereas a straighforward estimation of the class labels based on the minimum reconstruction error leads to a noisy segmentation (Fig.~\ref{fig:Im4_etapes}(b)). The label expansion algorithm via graph cut is crucial to create larger label support regions and suppress the majority of the small isolated label supports (Fig.~\ref{fig:Im4_etapes}(c)). Note that the graph cut algorithm does not take the label image in Fig.~\ref{fig:Im4_etapes}(b) as an input parameter but the data cost matrix consisting of the reconstruction errors $E(y,D^c)$  computed for each pixel and each class. The final erosion step erodes the last remaining small label supports and slightly erodes the edges in order to obtain a clean label image (Fig.~\ref{fig:Im4_etapes}(d)), close to the ground truth (Fig.~\ref{fig:Im4_etapes}(e)). The erosion algorithm takes the label image obtained after the graph cut algorithm (Fig.~\ref{fig:Im4_etapes}(c)) as an initial segmentation, and uses the same data cost matrix. 

%Image 4 aux différentes étapes (source, avant gc, après gc, finale, vérité terrain)

%Results

%Comparison to the state of the art
Finally, we compare in Table \ref{tab:all_res} the classification error rates of our method with several methods from the literature. In \cite{randen_filtering_1999}, introducing the dataset, numerous filtering methods are compared and the best result obtained for each test picture is presented.
The authors of \cite{maenpaa_texture_2000} have improved the previous results using the Local Binary Pattern operator on texture patches and by computing histograms of the values to characterize a texture.\footnote{We report the corrected results in an erratum posted by the authors at http://www.cse.oulu.fi/wsgi/CMV/SupervisedTextureSegmentation} A multi-scale version, considering several patch sizes, has also been studied.
The authors of\cite{di_lillo_texture_2007} have then proposed to extract texture discriminative features in the frequency domain by applying a Fourier transform in polar coordinates, followed by  dimensionality reduction via PCA (Principal Components Analysis) or the computation of Fisher coefficients. Centroids are then computed for each class with a vector quantization method.
The results presented in \cite{mairal_discriminative_2008} are also included in our comparison, using reconstructive (R) or discriminative (D) dictionaries, and a graph cut based smoothing method.
Finally, the results obtained with the $\alpha$-erosion method in \cite{skretting_energy_2014} are added. A dictionary is learnt per class with the RLS-DLA algorithm \cite{skretting_recursive_2010} and class labels are estimated based on the approximation errors computed for each pixel and each class in an energy minimization step. In this step, a Gaussian filter is applied before applying the $\alpha$-erosion algorithm, followed by further erosion of the edges of the label support regions in order to smooth their borders. In the smoothing steps of our method, the label expansion algorithm uses a random ordering of the labels to be expanded in each iteration. The classification results can thus change from one trial to another, despite the use of the same dictionaries and parameters. We thus perform the smoothing steps $20$ times and report the average error over these $20$ random trials. The difference between different trials remains small in general for the same image.

It is seen in Table \ref{tab:all_res} that, over the 12 different texture classification experiments, our method gives the best results in three experiments and is among the best two methods in nine experiments. Except for two problematic images (test images $5$ and $6$) the classification error of our method does not exceed that of the state of the art by more than $1\%$. Our average classification error over the $12$ experiments is $2.86\%$, which is the smallest among the compared methods.

%Tableau : résultats état de l'art et mes résultats (mean)
%[28],[17],[16],[R2],[D2],[Skretting],[Moi(mean)]
\begin{table}[!t]
%\begin{center}
\centering
\begin{tabular}{|c|c c c c c c|c|}
  \hline
  Im. & \cite{randen_filtering_1999}  & \cite{maenpaa_texture_2000} & \cite{di_lillo_texture_2007} & \cite{mairal_discriminative_2008}(R) & \cite{mairal_discriminative_2008}(D) & \cite{skretting_energy_2014} & Our meth. \\
  \hline
  1  & 7.2  & 7.5  & 3.37  & 1.69  & \textbf{1.61}  & 2.00 & \textbf{1.25} \\
  2  & 18.9 & 15.5 & 16.05 & 36.5  & 16.42 & \textbf{3.24} & \textbf{3.42} \\
  3  & 20.6 & 10.9 & 13.03 & 5.49  & 4.15  & \textbf{4.01} & \textbf{3.05} \\
  4  & 16.8 & 8.4  & 6.62  & 4.60  & 3.67  & \textbf{2.55} & \textbf{2.59} \\
  5  & 17.2 & 7.9  & 8.15  & \textbf{4.32}  & 4.58  & \textbf{1.26} & 6.60 \\
  6  & 34.7 & 16.1 & 18.66 & 15.50 & 9.04  & \textbf{6.72} & \textbf{8.20} \\
  7  & 41.7 & 20.3 & 21.67 & 21.89 & 8.80  & \textbf{4.14} & \textbf{2.36} \\
  8  & 32.3 & 16.2 & 21.96 & 11.80 & \textbf{2.24}  & 4.80 & \textbf{3.13} \\
  9  & 27.8 & 20.2 & 9.61  & 21.88 & \textbf{2.04}  & 3.90 & \textbf{2.06} \\
  10 & 0.7  & 0.3  & 0.36  & \textbf{0.17}  & \textbf{0.17}  & 0.42 & 0.23 \\
  11 & \textbf{0.2}  & 0.9  & 1.33  & 0.73  & 0.60  & 0.61 & \textbf{0.43} \\
  12 & 2.5  & 5.0  & 1.14  & \textbf{0.37}  & 0.78  & \textbf{0.70} & 0.94 \\
  \hline
  Av. & 18.4 & 10.8 & 10.16 & 10.41 & 4.50 & \textbf{2.87} & \textbf{2.86} \\
  \hline
  \end{tabular}
  %\end{center}
  \vspace{0.2cm}
  \caption[Classification error rates (in \%) of our method for the test images in comparison with several state of the art methods.]
   {Classification error rates (in \%) of our method for the test images in comparison with several methods from the state of the art. The best two results for each image are in bold.}
   \label{tab:all_res}
\end{table}

\begin{figure}[!htb]
  \centering
  \subfloat[Test image 7]{\includegraphics[trim=1.75cm 1.25cm 1.75cm 0.75cm,clip=true,scale=0.2]{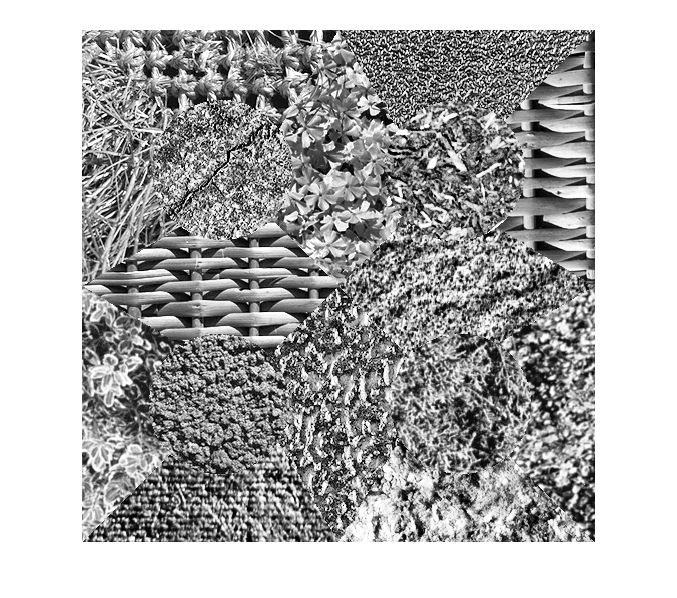}}
  %\hskip1em
  %\\
  \subfloat[Label image]{\includegraphics[trim=1.75cm 1.25cm 1.75cm 0.75cm,clip=true,scale=0.2]{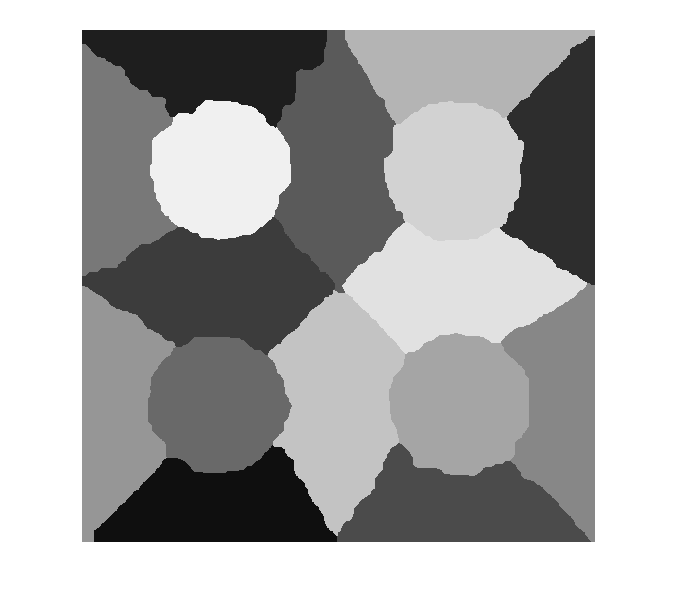}}
  %\\
  %\hskip1em
  \subfloat[Ground truth]{\includegraphics[trim=1.75cm 1.25cm 1.75cm 0.75cm,clip=true,scale=0.2]{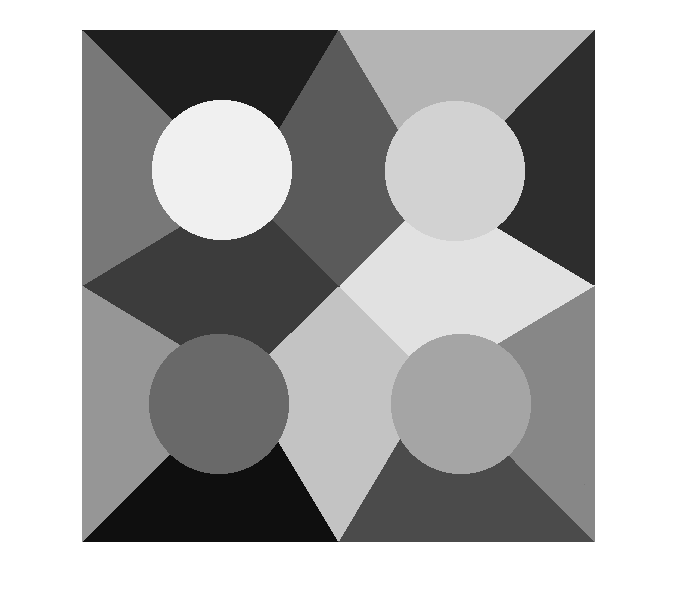}}
   \vspace{0cm}
  \caption[Test image 7 and its label image compared to the ground truth.]
  {Test image 7 and its label image compared to the ground truth.}
   \label{fig:Im7_res}
\end{figure}

\begin{figure}[!htb]
  \centering
  \subfloat[Test image 9]{\includegraphics[trim=1.75cm 1.25cm 1.75cm 0.75cm,clip=true,scale=0.2]{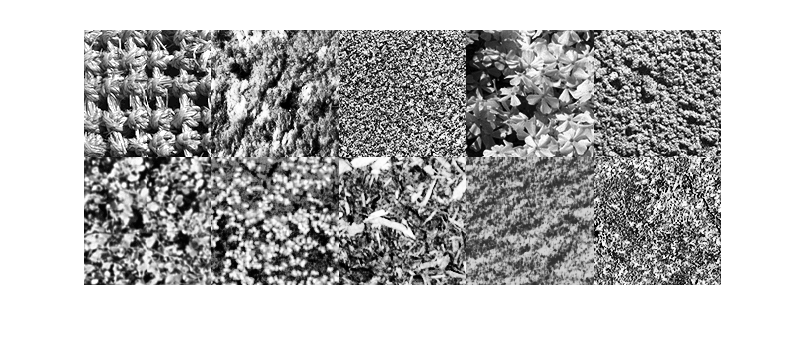}}
  %\hskip1em
  \\
  \subfloat[Label image]{\includegraphics[trim=1.75cm 1.25cm 1.75cm 0.75cm,clip=true,scale=0.2]{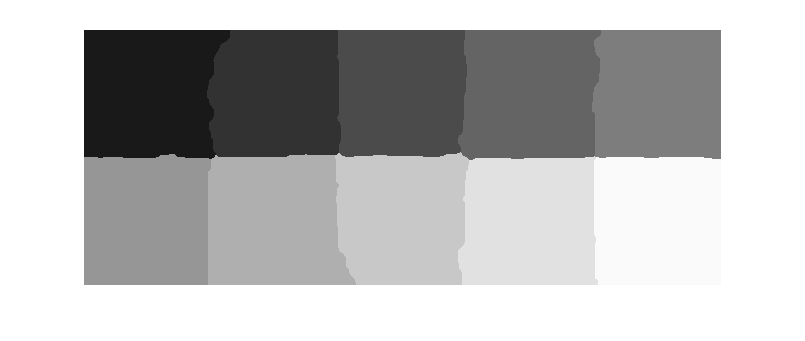}}
  %\\
  %\hskip1em
  \subfloat[Ground truth]{\includegraphics[trim=1.75cm 1.25cm 1.75cm 0.75cm,clip=true,scale=0.2]{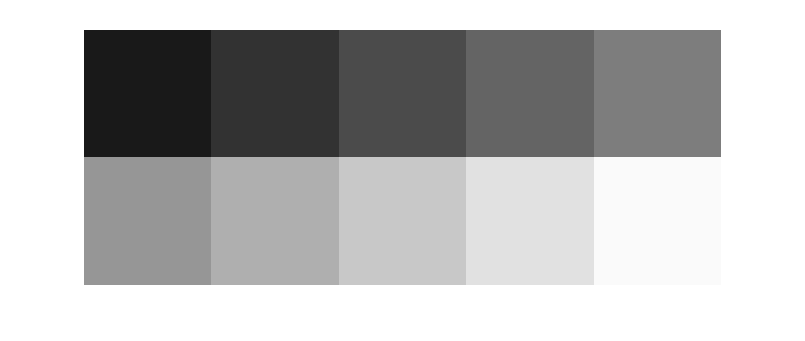}}
   \vspace{0cm}
  \caption[Test image 9 and its label image compared to the ground truth.]
  {Test image 9 and its label image compared to the ground truth.}
   \label{fig:Im9_res}
\end{figure}

\begin{figure}[!htb]
  \centering
  \subfloat[Test image 11]{\includegraphics[trim=1.75cm 1.25cm 1.75cm 0.75cm,clip=true,scale=0.2]{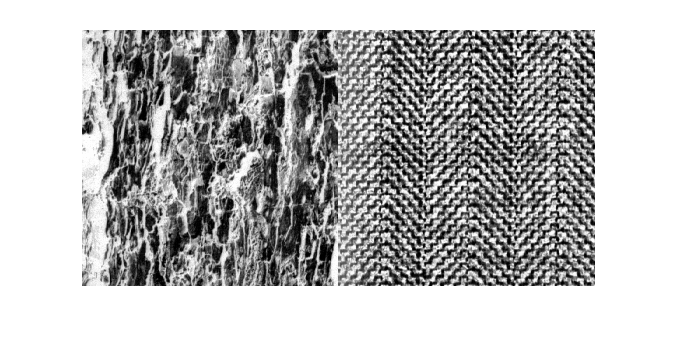}}
  %\hskip1em
  \subfloat[Label image]{\includegraphics[trim=1.75cm 1.25cm 1.75cm 0.75cm,clip=true,scale=0.2]{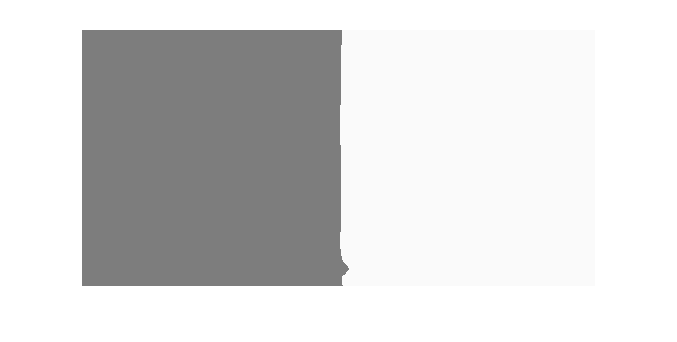}}
  %\\
  %\hskip1em
  \subfloat[Ground truth]{\includegraphics[trim=1.75cm 1.25cm 1.75cm 0.75cm,clip=true,scale=0.2]{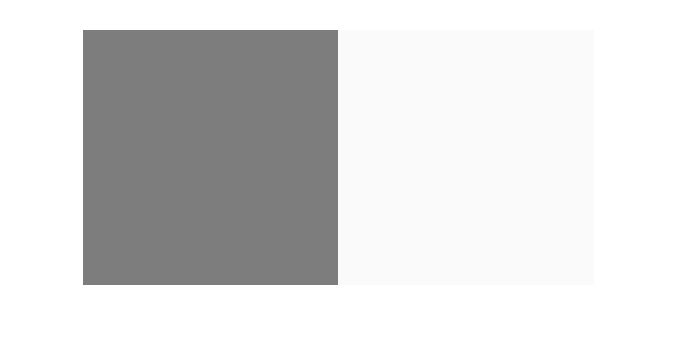}}
   \vspace{0cm}
  \caption[Test image 11 and its label image compared to the ground truth.]
  {Test image 11 and its label image compared to the ground truth.}
   \label{fig:Im11_res}
\end{figure}

% Cas positifs
Some example classification results are presented for several test images in Figures \ref{fig:Im7_res}, \ref{fig:Im9_res}, and \ref{fig:Im11_res}. We observe that the only zones of misclassification are concentrated within a thin band over the edges between label supports, and the classification performance of our method is quite satisfactory in these experiments.

% Image 7

% Image 9

% Image 11

\begin{figure}[!htb]
  \centering
  \subfloat[Test image 5]{\includegraphics[trim=1.75cm 1.25cm 1.75cm 0.75cm,clip=true,scale=0.3]{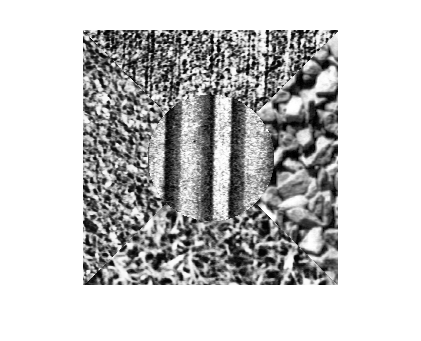}}
  \hskip1em
  \subfloat[Min error]{\includegraphics[trim=1.75cm 1.25cm 1.75cm 0.75cm,clip=true,scale=0.3]{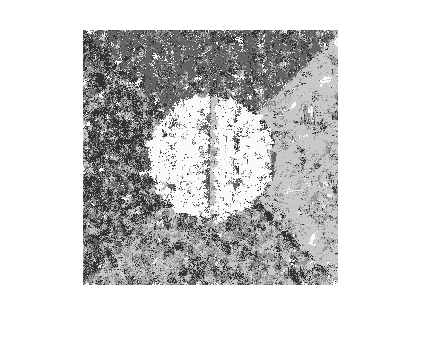}}
  \\
  \subfloat[After graph cut]{\includegraphics[trim=1.75cm 1.25cm 1.75cm 0.75cm,clip=true,scale=0.3]{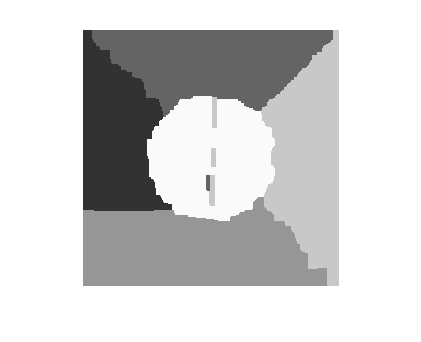}}
  %\hskip1em
  \subfloat[After erosion]{\includegraphics[trim=1.75cm 1.25cm 1.75cm 0.75cm,clip=true,scale=0.3]{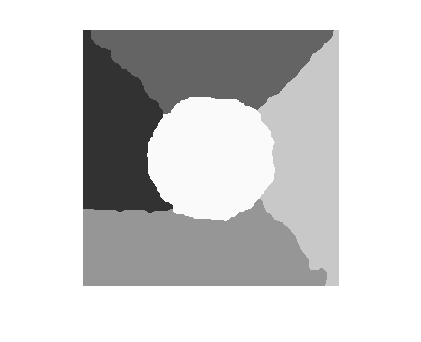}}
  %\hskip1em
  \subfloat[Ground truth]{\includegraphics[trim=1.75cm 1.25cm 1.75cm 0.75cm,clip=true,scale=0.3]{fig/5cl_gt.png}}
  \\
   \vspace{0cm}
  \caption[Test image 5 and the successive steps of the classification algorithm.]
  {Test image 5 and the successive steps of the classification algorithm.}
   \label{fig:Im5_res}
\end{figure}

\begin{figure}[!htb]
  \centering
  \subfloat[Test image 6]{\includegraphics[trim=1.75cm 1.25cm 1.75cm 0.75cm,clip=true,scale=0.2]{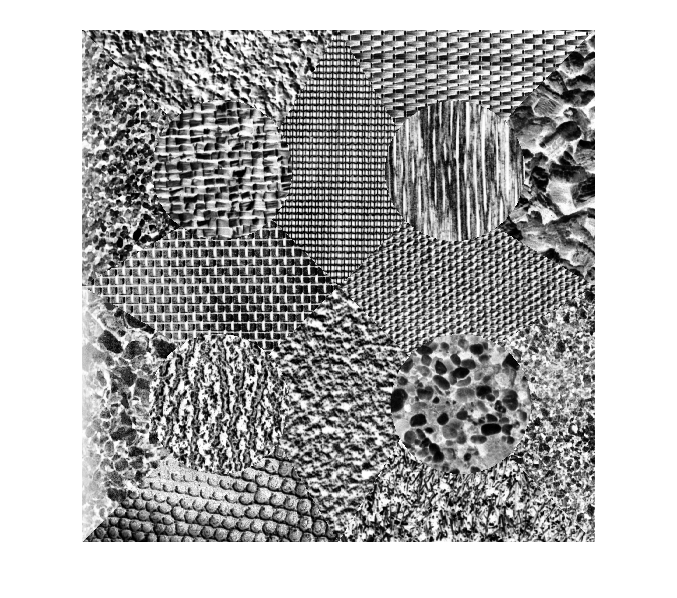}}
  %\hskip1em
  \subfloat[Min error]{\includegraphics[trim=1.75cm 1.25cm 1.75cm 0.75cm,clip=true,scale=0.2]{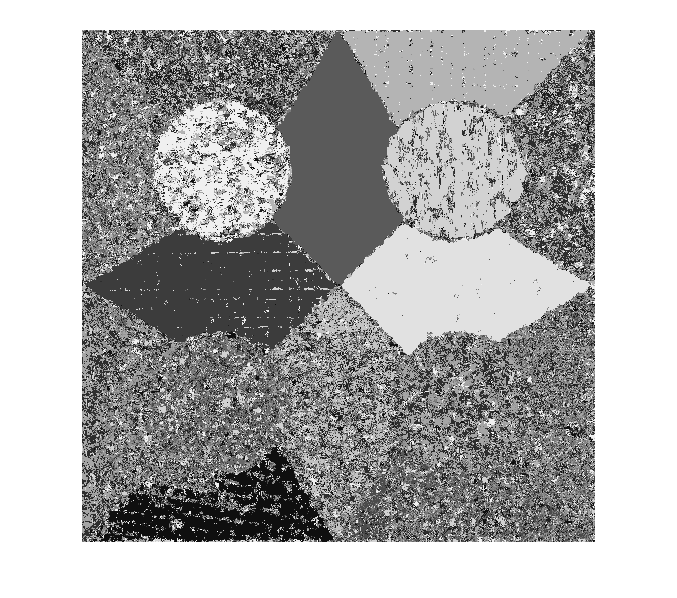}}
  \\
  %\hskip1em
  \subfloat[After graph cut]{\includegraphics[trim=1.75cm 1.25cm 1.75cm 0.75cm,clip=true,scale=0.2]{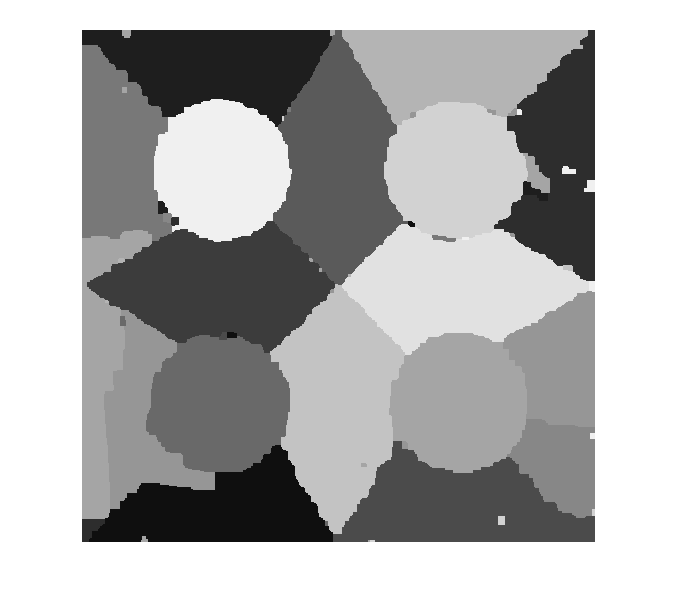}}
  \subfloat[After erosion]{\includegraphics[trim=1.75cm 1.25cm 1.75cm 0.75cm,clip=true,scale=0.2]{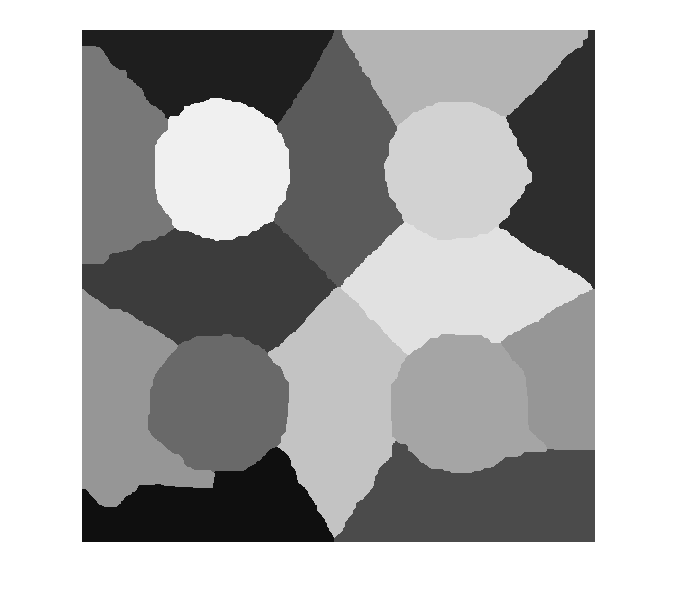}}
  \subfloat[Ground truth]{\includegraphics[trim=1.75cm 1.25cm 1.75cm 0.75cm,clip=true,scale=0.2]{fig/16cl_gt.png}}
   \vspace{0cm}
  \caption[Test image 6 and the successive steps of the classification algorithm.]
  {Test image 6 and the successive steps of the classification algorithm.}
   \label{fig:Im6_res}
\end{figure}

% Cas problématiques
Meanwhile, the texture classification experiments corresponding to the test images 5 and 6 are particularly challenging and these are the only settings where our method gives a classification error rate superior to $4\%$ (Table \ref{tab:all_res}). Our results on these two test images are presented in Figures \ref{fig:Im5_res} and \ref{fig:Im6_res}. For the test image $5$ in Fig.~\ref{fig:Im5_res}, the main factor increasing the classification error is the misclassification of the region in the bottom-left corner of the picture, where the label of the bottom texture spreads too much on the leftmost texture. Indeed, the border between these two textures on the test image is difficult to see even for the human eye and the two texture classes have very similar characteristics. Hence, the erroneous label can easily be diffused over the leftmost texture and it is not surprising to observe a relatively high misclassification rate in this experiment.

For the test image 6 given in Fig.~\ref{fig:Im6_res}, the problem is different. When we observe the final label image (Fig.~\ref{fig:Im6_res}(d)), the major regions of misclassification are over the textures on the left at the bottom, where the label spreads too much, and on the right at the bottom, where the whole texture has a wrong label. In the label image obtained after the graph cut based smoothing step (Fig.~\ref{fig:Im6_res}(c)), we notice that the misclassification regions for these two textures lie respectively  on the left part of the first one and at the top of the second one. When we look at these specific areas in the original image (Fig.~\ref{fig:Im6_res}(a)), we can see that they seem over-exposed and thus brighter in comparison to the rest of the textures. Meanwhile, this over-exposure is not present in the training images, which can disturb the classification algorithm as this kind of variation has not been learnt, and lead to confusion with the other classes.

We also notice that textures with regular and small patterns are easily classified, even without the smoothing steps (Fig.~\ref{fig:Im6_res}(b)), as the learning is easier.

%Image 5

% Image 6

\subsection{Enriching the training dataset}

%Over-exposure results

In order to deal with the over-exposure problems, particularly present on the test image $6$, we propose to enrich the training dataset by adding some over-exposed versions of the training images.
An over-exposed image, called $I_{exp}$, is created from the training image $I$ with the following equations
\begin{equation}
I_{tmp} = I + \Delta_{exp}
\end{equation}
\begin{equation}
M = \max(I_{tmp})
\end{equation}
\begin{equation}
m = \min(I)
\end{equation}
\begin{equation}
I_{exp} = round\left(  \frac{255 \ ( I_{tmp} - m)}{M-m}\right)
\end{equation}
where $\Delta_{exp}$ is the exposure offset added to the image $I$, $M$ is the maximum value in $I_{tmp}$ after the exposure offset $\Delta_{exp}$ has been added to the image $I$, and $m$ is the minimum value in $I$. 

In this way, the training set corresponding to the test image $6$ is augmented by generating different over-exposed versions of the training images with over-exposure levels $\Delta_{exp} = 100, 300, 500$, and $700$. We  balance the number of original and over-exposed training samples by generating a total of $40000$ over-exposed samples ($10000$ samples for each $\Delta_{exp}$ value) added to $40000$ original training samples. Dictionaries are then learnt from this new training dataset for the classification problem $6$.

%Thus, among a total of $80000$ training vector per class, the half are over-exposed 

\begin{figure}[!htb]
  \centering
  \subfloat[After graph cut]{\includegraphics[trim=1.75cm 1.25cm 1.75cm 0.75cm,clip=true,scale=0.22]{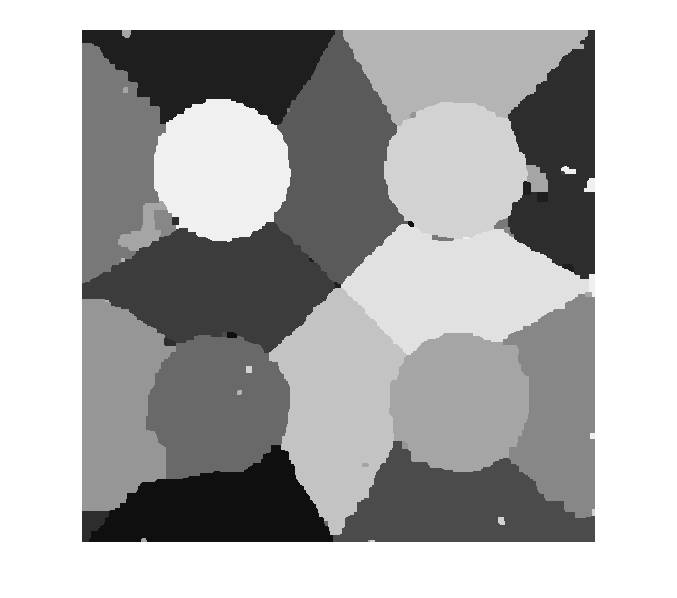}}
  \hskip2em
  \subfloat[After erosion]{\includegraphics[trim=1.75cm 1.25cm 1.75cm 0.75cm,clip=true,scale=0.22]{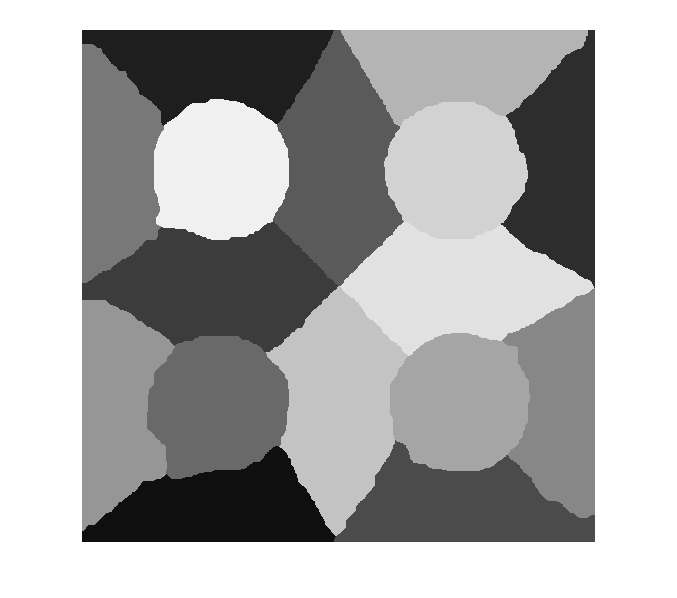}}
  \\
  %\hskip1em
%  \subfloat[After graph-cut (case 2)]{\includegraphics[trim=1.75cm 1.25cm 1.75cm 0.75cm,clip=true,scale=0.22]{fig/Im6_b_Expo2_ex7,50.png}}
%  \hskip2em
%  \subfloat[After erosion (case 2)]{\includegraphics[trim=1.75cm 1.25cm 1.75cm 0.75cm,clip=true,scale=0.22]{fig/Im6_c_Expo2_ex7,50.png}}
%  \subfloat[Vérité terrain]{\includegraphics[trim=1.75cm 1.25cm 1.75cm 0.75cm,clip=true,scale=0.3]{chapitre6/fig/16cl_gt.png}}
   \vspace{0cm}
  \caption[Label images obtained for the test image $6$ after enriching the training dataset with over-exposed patches.]
  {Label images obtained for the test image $6$ after enriching the training dataset with over-exposed patches.}
   \label{fig:Im6_res_Expo2}
\end{figure}

Some classification results obtained for the test image are presented in Fig.~\ref{fig:Im6_res_Expo2}. It can be observed that augmenting the training data set with over-exposed samples has the potential to improve the classification performance. However, we have also observed that in the smoothing step, the expansion of the labels with a random ordering in the graph cut method may produce a more erroneous label image in some random realizations of the experiment. We have obtained an average classification error of $4.36\%$ over 20 different random repetitions of the same experiment, whereas the error rate was $8.20\%$ before enriching the learning dataset for the test image 6. If we take into account this new error rate for the test image $6$, the mean classification error rate computed over the $12$ experiments is reduced to $2.54\%$ from its previous value $2.86\%$ in Table \ref{tab:all_res}.

%The classification results obtained for the test image $6$ typically change between two cases, according to the random order of the labels during the label expansion in the graph-cut method used for the smoothing (Fig.\ref{fig:Im6_res_Expo2}). Case $1$ is the favorable case: the over-exposure problem is solved and the label image presents a classification error rate of $2.35\%$ on Figure \ref{fig:Im6_res_Expo2}(b). For case $2$, the problem is solved for the texture on the left at the bottom, but not for the one on the right at the bottom, whose label is entirely wrong, giving a classification error rate of $7.50\%$ on Figure \ref{fig:Im6_res_Expo2}(d). By looking at the right-bottom label on the label image after the graph-cut-based smoothing for case $2$ (Fig.\ref{fig:Im6_res_Expo2}(c)), we can observe that only the over-exposed part of the texture at the center is still misclassified, the rest of the texture having the right label. This is then the erosion step that affects the wrong label to the whole texture.

% Image 6 Expo2

%When the smoothing steps are repeated $20$ times to obtain $20$ classification results, the classification error rates are between $2.17\%$ and $7.65\%$. The mean error is $4.36\%$ whereas it was $8.20\%$ before enriching the learning dataset. If we take into account this new value for the test image $6$ (without changing the values for the other test images), the mean classification error rate computed on the $12$ images falls to $2.54\%$.

Enriching the training dataset has thus improved the results. The new over-exposed training data have been helpful for learning dictionary structures containing more information and more conscious of this possible intra-class exposure variation. This solution could be applied to other test images as well undergoing the same problem.

\section{Conclusion}
\label{sec:conclusion}

In this paper, we have proposed a method for learning discriminative multilevel structured dictionaries for supervised image classification. We have presented a classification algorithm that learns one dictionary per class, where test images are classified with respect to their reconstruction errors on these dictionaries. For the construction of the dictionaries, we have adopted the Adaptive Structure derived from a tree structure, which we made discriminant with a novel objective function to learn multilevel dictionaries that are both reconstructive and discriminative. The proposed dictionaries thus have a high learning capacity due to their multilevel topology and are well-adapted to the classification of images with high intra-class variation.  An affinity matrix has been incorporated in the objective function to adjust the discrimination of a class from the others depending on their pairwise affinities. A combination of two smoothing methods has been used to obtain a clean segmentation and classification of the textures in the test image. Experiments conducted on a common dataset of texture images have shown competitive results with the state of the art. We have finally proposed to enrich the training dataset to deal with over-exposure problems.

Enriching the dataset seems promising and future efforts may focus on more complex and realistic over-exposure models. Applying discrimination to all the dictionaries in the multilevel structure may also potentially be of interest, but might increase the complexity of the learning. Finally, a last future direction is to explore other affinity measures in the construction of the affinity matrix, in order to better characterize the pairwise similarities of classes and thus enhance the discrimination capability of the learnt dictionaries.

% or


%\appendices
%\section{Proof of the minimization to update an atom}
% \appendix[Proof of the minimization to update an atom]
% Appendix text goes here.

%\section{}

\section*{Acknowledgment}
This work was supported by Airbus Defence \& Space.

\ifCLASSOPTIONcaptionsoff
  \newpage
\fi



\bibliographystyle{IEEEtran}
\bibliography{biblio}
\end{document}